\documentclass[11pt]{amsart}

\usepackage{colortbl,xcolor}
\usepackage[vmargin = 1in, hmargin = 1in]{geometry}
\usepackage{siunitx}
\usepackage{paralist}
\usepackage{graphicx}
\usepackage[colorlinks=true, linkcolor=blue, citecolor=blue]{hyperref}
\usepackage{booktabs}

\definecolor{red}{RGB}{200,16,46}

\newcolumntype{R}{>{\columncolor{gray!40}}r}
\newcolumntype{L}{>{\columncolor{gray!40}}l}
\newcolumntype{C}{>{\columncolor{gray!40}}c}

\newcommand{\fun}[1]{\operatorname{#1}}
\newcommand{\defeq}{\ensuremath{\mathrel{\mathop:}=}}
\newcommand{\eqdef}{\ensuremath{=\mathrel{\mathop:}}}
\newcommand{\ns}[1]{\ensuremath{\mathbb{#1}}}
\DeclareMathOperator*{\argmin}{argmin}
\DeclareMathOperator*{\argmax}{argmax}
\DeclareMathOperator*{\minopt}{minimize}
\DeclareMathOperator*{\card}{card}
\newcommand{\secref}[1]{Sec.~\ref{#1}}
\newcommand{\tabref}[1]{Tab.~\ref{#1}}
\newcommand{\figref}[1]{Fig.~\ref{#1}}
\newcommand{\iquote}[1]{``\textit{#1}''}
\newcommand{\mli}[1]{\mathit{#1}}

\title[Stochastic Neural Networks for Cell Tracking]{Stochastic Neural Networks for Automatic Cell Tracking in Microscopy Image Sequences of Bacterial Colonies}

\author[S. Sarmadi et al.]{Sorena Sarmadi$^{1}$, James J. Winkle$^{1}$, Razan N. Alnahhas$^{2}$, Matthew R. Bennett$^{3}$, Kre\v{s}imir Josi\'{c}$^{1,4}$, Andreas Mang$^{1}$ and Robert Azencott$^{1}$}

\thanks{$^{1}$ Department of Mathematics, University of Houston; \{\texttt{andreas},\texttt{josic},\texttt{razencot}\}\texttt{@math.uh.edu}\\
$^{2}$ Department of Biomedical Engineering, Boston University\\
$^{3}$ Departments of Biosciences and Bioengineering, Rice University; \texttt{matthew.bennett@rice.edu}\\
$^{4}$ Department of Biology and Biochemistry, University of Houston
}

\begin{document}

\maketitle

\begin{abstract}
Our work targets automated analysis to quantify the growth dynamics of a population of bacilliform bacteria. We propose an innovative approach to frame-sequence tracking of deformable-cell motion by the automated minimization of a new, specific cost functional. This minimization is implemented by dedicated Boltzmann machines (stochastic recurrent neural networks). Automated detection of cell divisions is handled similarly by successive minimizations of two cost functions, alternating the identification of children pairs and parent identification. We validate the proposed automatic cell tracking algorithm using recordings of simulated cell colonies that closely mimic the growth dynamics of \emph{E. coli} in microfluidic traps and real data. On a batch of 1100 simulated image frames, cell registration accuracies per frame ranged from 94.5\% to 100\%, with a high average. Our initial tests using experimental image sequences (i.e., real data) of \emph{E. coli} colonies also yield convincing results, with a registration accuracy ranging from 90\% to 100\%.
\end{abstract}

\section{Introduction}\label{s:intro}

Technology advances have led to increasing magnitudes of data generation with increasing levels of precision~\cite{ButtsWilmsmeyer:2020a,Sivarajah:2017a}. However, data generation presently far outpaces data analysis and drives the requirement for analyzing such large-scale data sets with automated tools~\cite{Balomenos:2017a,Klein:2012a,Stylianidou:2016a}. The main goal of the present work is to develop computational methods for an automated analysis of microscopy image sequences of colonies of \emph{E.\ coli} growing in a single layer. Such recordings can be obtained from colonies growing in microfluidic devices, and they provide a detailed view of individual cell-growth dynamics as well as population-level, inter-cellular mechanical and chemical interactions~\cite{Bennett:2009a,Danino:2010a,Mather:2010a}.

However, to understand both variability and lineage-based correlations in cellular response to environmental factors and signals from other cells requires the tracking of large numbers of individual cells across many generations. This can be challenging, as large cell numbers tightly packed in microfluidic devices can compromise spatial resolution, and toxicity effects can place limits on the temporal resolution of the recordings~\cite{ElNajjar:2020a,Icha:2017a}. One approach to better understand and control the behavior of these bacterial colonies is to develop computational methods that capture the dynamics of gene networks within single cells~\cite{Bennett:2009a,Kim:2019a,Winkle:2017a}. For these methods to have a practical impact, one ultimately has to fit the models to the data, which allows us to infer hidden parameters (i.e., characteristics of the behavior of cells that cannot be measured directly). Image analysis and pattern recognition techniques for biological imaging data~\cite{Carpenter:2006a,Kamentsky:2011a,McQuin:2018a}, like the methods developed in the present work, can be used to track lineages and thus automatically infer how gene expression varies over time. These methods can serve as an indispensable tool to extract information to fit and validate both coarse and detailed models of bacterial population, thus allowing us to infer model parameters from recordings.

\begin{figure}
\centering
\includegraphics[width=0.7\textwidth]{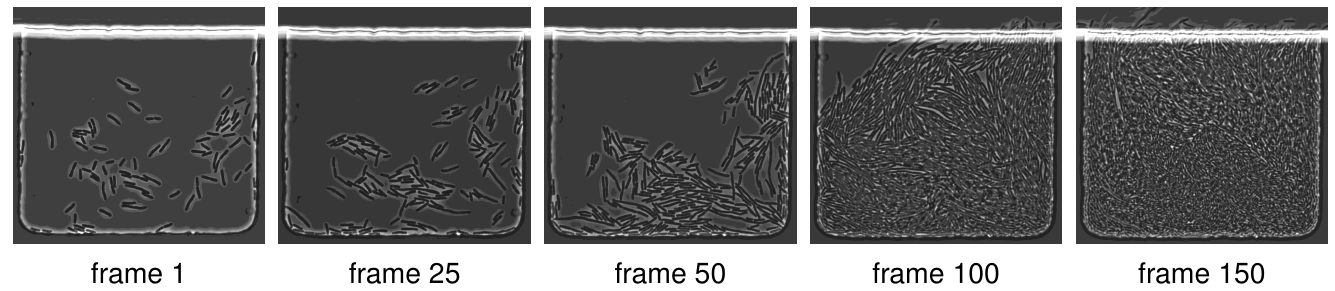}
\caption{\textit{\textbf{Typical microscopy image sequence}}. We show five frames out of a total of 150 frames of an image sequence showing the dynamics of \emph{E.\ coli} in a microfluidic device~\cite{Alnahhas:2019a} (real laboratory image data). These cells are are about 1~$\mu$m in diameter and on average 3~$\mu$m in length, and they divide about every 30 minutes. The original images exported from the microscope are 0.11~$\mu$m/pixel. We report results for these real datasets in \secref{s:map_split}.\label{f:exemplary-data}}
\end{figure}

Here we describe an algorithm that provides \emph{quantitative} information about the population dynamics, including the life cycle and lineage of cells within a population from recordings of cells growing in a mono-layer. A typical sequence of frames of cells growing in a microfluidic trap is shown in \figref{f:exemplary-data}. We describe the design and validation of algorithms for tracking individual cells in sequences of such images~\cite{Alnahhas:2020a,Kim:2019a,Locke:2009a}. After segmentation of individual image frames to identify each cell, tracking individual cells from frame to frame is a combinatorial problem. To solve this problem we take into account the unknown cell growth, cell motion, and cell divisions that occur between frames. Segmentation and tracking are complicated by imaging noise and artifacts, overlap of bacteria, similarity of important cell characteristics across the population (shape; length; and diameter), tight packing of bacteria, and large interframe durations resulting in significant cell motion, and up to a 30\% increase in individual cell volume.

\subsection{Related Work}\label{s:related}

The present work focusses on tracking \emph{E.\ coli} in time series of images. A comparison of different cell-tracking algorithms can be found in~\cite{Hand:2009a,Ulman:2017a}. Multi-object tracking in video sequences and object recognition in time series of images is a challenging task that arises in numerous applications in computer vision~\cite{MarvastiZadeh:2021a,Yilmaz:2006a}. In(biomedical) image processing, motion tracking is often referred to as \iquote{image registration}~\cite{Hand:2009a,Lucas:1981a,Mang:2015a,Mang:2017a,Mang:2019a} or \iquote{optical flow}~\cite{Borzi:2002b,Horn:1981a,Delpiano:2012a,Madrigal:2015a}. Typical solutions used in the defense industry, for instance, track \emph{small numbers} of fast moving targets by image sequence analysis at pixel levels and use sophisticated reconstruction of the optical flow, combined with real time segmentation, and quick combinatoric exploration at each image frame. Initially, we did implement several well known algorithms for reconstruction of the optical flow but the results we obtained were not satisfactory due to long interframe times and high noise levels. Moreover, we are not interested in tracking individual pixels but rather cells (i.e., rod-shaped, deformable shapes), while recognizing events of cell division and recording cell lineage. Consequently, we decided to first segment each image frame to isolate each cell, and then to match cells between successive frames.

As for the  problem at hand, one approach proposed in prior work to simplify the tracking task is to make the experimental setup more rigid by confining individual cell lineages to small tubes; the associated microfluidic device is called a \iquote{mother machine}~\cite{Banerjee:2020a,Jug:2014a,Lugagne:2020a,Ollion:2019a,Sauls:2019a,Smith:2019a}. The microfluidic devices we consider here yield more complicated data as cells are allowed to move and multiply freely in two dimensions (constrained to a mono-layer). We refer to \figref{f:exemplary-data} for a typical sequence of experimental images considered in the present work.

Turning to methods that work on more complex biological cell imaging data, we can distinguish different classes of tracking methods. \iquote{Model-based evolution methods} operate on the image intensities directly. They rely on particle filters~\cite{Arbelle:2018a,Okuma:2004a,Smal:2006a} or active contour models~\cite{Kervrann:2002a,Li:2008a,Wang:2007a,Yang:2005a,Sethuraman:2012a}. These methods work well if the cells are not tightly packed. However, they may lead to erroneous results if the cells are close together, the inter-cellular boundaries are blurry, or the cells move significantly. Our work belongs to another class---the so called \iquote{detection-and-association methods}~\cite{Balomenos:2015a,Bise:2011a,Bise:2009a,Kanade:2011a,Primet:2008a,Ronneberger:2015a,Su:2013a,Wang:2010a,Jiuqing:2017a,Zhou:2019a,Sixta:2020a}, which first detect cells in each frame and then solve the tracking problem/association task across successive frames. (We note that the segmentation and tracking of cells does not necessarily need to be implemented in two distinct steps. In many image sequence analyses, implementing these two steps jointly can be beneficial~\cite{Arbelle:2018a,Hayashida:2020a,Payer:2018a,Primet:2008a,Payer:2019a,Zhou:2019a}. However, for the clarity of exposition and easier implementation of our new tracking technique, we present these steps separately.) Doing so necessitates the segmentation of cells within individual frames. We refer to~\cite{Vicar:2019a} for an overview of cell segmentation approaches. Deep learning strategies have been widely used for this task~\cite{AlKofahi:2018a,Falk:2019a,Lux:2019a,Moen:2019b,Payer:2019a,Ronneberger:2015a,Rempfler:2018a,Stringer:2021a,Stylianidou:2016a,Zhou:2019a}. We consider a framework based on convolutional neural networks ({\bf CNN}s). Others have also used CNNs for cell segmentation~\cite{Akram:2016a,Lux:2019a,Nishimura:2020a,Rempfler:2017a,Rempfler:2018a}. We omit a detailed discussion of our segmentation approach within the main body of this paper, as we do not view it as our main contribution (see \secref{s:contributions}). However, the interested reader is referred to \secref{s:cell-segmentation} for some insights. To solve the tracking problem after the cell detection, many of the methods cited above use hand-crafted association scores based on the proximity of the cells and shape similarity measures~\cite{Kanade:2011a,Bise:2011a,Su:2013a,Zhou:2019a}. We follow this approach here. We note that we not only consider local association scores between cells but also include measures for the integrity of a cell's neighborhood (i.e., \iquote{context information}).

Our method is tailored for tracking cells in tightly packed colonies of rod-shaped \emph{E.\ coli} bacteria. This problem has been considered previously~\cite{Balomenos:2015a,Primet:2008a,Stylianidou:2016a,Wang:2010a}. However, we are not aware of any large-scale datasets that provide ground truth tracking data for these types of recordings, but note that there are community efforts for providing a framework for testing cell tracking algorithms~\cite{Maska:2014a,Ulman:2017a} (see, e.g.,~\cite{Arbelle:2018a,Loffler:2021a}).\footnote{Cell tracking challenge: \url{http://celltrackingchallenge.net} (accessed 03/2021).} Works that consider these data are for example~\cite{Arbelle:2018a,Lux:2019a,Nishimura:2020a,Payer:2018a,Payer:2019a,Zhou:2019a}. The cells in this dataset have significantly different characteristics compared to those considered in the present work. As we describe below, our approach is based on distinct characteristics of the bacteria cells and, consequently, does not directly apply to these data. Therefore, we have developed our own validation and calibration framework (see \secref{s:synthetic}).

Standard graph matching algorithms (see, e.g., \cite{Vo:2008a}) do not directly apply to our problem. Indeed, a fundamental complication is that cells can divide between successive images. Hence, each assignment from one frame to its successor is not a one-to-one mapping but a one-to-many correspondence. More advanced graph matching strategies are described in~\cite{Pierskalla:1968a,Gilbert:1988a}. Graph-based matching strategies for cell-tracking that are somewhat related to our approach are described in~\cite{Chakraborty:2015a,Liu:2010a,Liu:2011a,Loffler:2021a,Liu:2018a}. Like the methods mentioned above, they consider various association scores for tracking. Individual cells are represented as nodes, and neighbors are connected through edges. Our approach also introduces cost terms for structural matching of local neighborhoods by specific scoring for single nodes, pairs of nodes, and triplets of nodes, after a (modified) Delaunay triangulation. By using a graph-like structure, cell divisions can be identified by detecting changes in the topology of the graph~\cite{Liu:2010a,Liu:2011a}. We tested a similar strategy, but came to the conclusion that we cannot reliably construct neighborhood networks between frames for which topology changes only occur due to cell division; the main issue we observed is that the significant motion of cells between frames can introduce additional topology changes in our neighborhood structure. Consequently, we decided to relax these assumptions.

\cite{Vo:2008a,Vo:2019a,Punchihewa:2018a,Kim:2017a} implement multi-target tracking in videos by stochastic models based on random finite set densities and variants thereof. The fit to the data is based on Gibbs sampling to maximize the posterior likelihood. A key challenge of these approaches is the estimation of an adequate finite number of Gibbs sampling iterations when one computes posterior distributions. Most Gibbs samplers are ergodic Markov chains on a finite but huge state spaces, so that their natural exponential rate of convergence is not a practically reassuring feature.

As mentioned above, some recent works jointly solve the tracking and segmentation problem~\cite{Arbelle:2018a,Hayashida:2020a,Payer:2018a,Primet:2008a,Payer:2019a,Zhou:2019a}. Contrary to observations we have made in our data, these approaches rely (with the exception of~\cite{Primet:2008a}) on the fact that the tracking problem is inherent to the segmentation problem (\iquote{tracking-by-detection methods}~\cite{Zhou:2019a}; see also \cite{Stylianidou:2016a}). That is, the key assumption made by many of these algorithms is that cells belonging to the same lineage overlap across frames (see also~\cite{Bise:2009a}). In this case, cell-overlap can serve as a good proxy for cell-tracking~\cite{Zhou:2019a}. We note that in our data we cannot guarantee that the frame rate is sufficiently high for this assumption to hold.

\cite{Payer:2018a,Nishimura:2020a,Hayashida:2020a} exploited machine learning techniques for segmentation \emph{and} motion tracking. One key challenge here is to provide adequate training data for these methods to be successful. Here, we describe simulation-based techniques that can be extended to produce training data, which we use for parameter tuning~\cite{Winkle:2017a,Winkle:2021a}.

The works that are most similar to ours are~\cite{Balomenos:2015a,Primet:2008a,Wang:2010a}. They perform a local search to identify the best cell-tracking candidates across frames. One key difference across these works are the matching criteria. Moreover, \cite{Balomenos:2015a,Primet:2008a} employ a local greedy-search, whereas we consider stochastic neural network dynamics for optimization. \cite{Wang:2010a} construct score matrices within a score based neighborhood tracking method; an integer programming method is used to generate frame-to-frame correspondences between cells and the lineage map. Other approaches that consider linear programming to maximize an association score function for cell tracking can be found in~\cite{Bise:2015a,Bise:2009a,Zhou:2019a}.

As we have mentioned in the abstract we obtain a tracking accuracy that ranges from 90\% to 100\%, respectively. Overall, our method is competitive with existing approaches: For example, \cite{Balomenos:2015a} report a tracking accuracy of up to 97\% for data that is similar to ours, while \cite{Chakraborty:2015a} report a tracking accuracy (spatial, temporal, and cell division detection) at the order of 95\% (between about 93\% and 98\%, respectively). The second group also reports results for their prior approach~\cite{Liu:2010a}, with an accuracy at the order of 90\% (ranging from about 87\% to 92\%, respectively). Accuracies reported in~\cite{Liu:2018a} range from about 92\% to 97\%, respectively. This work also includes a comparison to one of their earlier approaches~\cite{Liu:2011a} with an accuracy of up to 85\% and 89\% if the datasets are pre-aligned. We note that the data considered in \cite{Chakraborty:2015a,Liu:2018a,Liu:2011a,Liu:2010a} is quite different from ours. \cite{Arbelle:2018a,Loffler:2021a,Payer:2019a,Payer:2018a,Zhou:2019a} consider the data from the cell tracking challenge~\cite{Maska:2014a,Ulman:2017a} to evaluate the performance of their methods. As in the previously mentioned work, this data is again quite different from ours. To evaluate the performance of the methodology the so called \emph{acyclic oriented graph matching measure}~\cite{Matula:2015a} is considered. We refer to the webpage of the cell tracking challenge for details on the evaluation metrics (see \url{http://celltrackingchallenge.net/evaluation-methodology}). Based on these, the reported tracking scores are between 0.873 and 0.902 \cite{Arbelle:2018a}, 0.901 and 1.00 \cite{Loffler:2021a}, 0.950 and 0.987 \cite{Zhou:2019a}, 0.788 and 0.982 \cite{Payer:2019a} and 0.765 and 0.915 \cite{Payer:2018a} depending on the considered data set, respectively.

\subsection{Contributions}\label{s:contributions}

For image segmentation, we first apply two well-known, powerful variational segmentation algorithms to generate a large training set of correctly delineated single cells. We can then train a CNN dedicated to segmenting out each single cell. Using a CNN significantly reduces the runtime of our computational framework for cell identification. The frame-to-frame tracking of individual cells in tightly packed colonies is a significantly more challenging task, and is hence the main topic discussed in the present work. We develop a set of innovative automatic cell tracking algorithms based on the successive minimization of three dedicated  cost functionals. For each pair of successive image frames, minimizing these cost functionals over all potential cell registration mappings poses significant computational and mathematical challenges. Standard gradient descent  algorithms are inefficient for these discrete and highly combinatorial minimization problems. Instead, we implement the stochastic neural network dynamics of BMs, with architectures and energy functions tailored to effectively solve our combinatorial tracking problem. Our major contributions are:
\begin{inparaenum}[\bf i)]
\item The design of a multi-stage cell tracking algorithm, that starts with  a parent-children pairing step, followed by removal of identified parent-children triplets, and concludes with a cell to cell registration step.
\item The design of dedicated BM architectures, with several energy functions, respectively, minimized by true parent-children pairing and by true cell-to-cell registration. Energy minimizations are then implemented by simulation of BM stochastic dynamics.
\item The development of automatic algorithms for the estimation  of unknown  weight parameters of our BM energy functions, using convex-concave programming tools~\cite{Agrawal:2018a,Diamond:2016a,Shen:2016a}.
\item The evaluation of our methodology on synthetic and real image sequences of cell colonies. The massive effort involved in human expert  annotation of  cell colony recordings limits the availability of \iquote{ground truth tracking} data for dense bacterial colonies. We therefore first validated the accuracy of our cell tracking algorithms on recordings of simulated cell colonies, generated by the dedicated cell colony simulation software~\cite{Winkle:2017a,Winkle:2021a}. This provided us with ground truth frame-by-frame registration for cell lineages, enabling us to validate our methodology.
\end{inparaenum}

\subsection{Outline}\label{s:outline}

In \secref{s:data} we describe the synthetic image sequence (see \secref{s:synthetic}) and experimental data (see \secref{s:real-data}) of cell colonies considered as benchmarks for our cell tracking algorithms. In \secref{s:cell_characteristics} we describe key cell characteristics considered in our tracking methodology to define metrics that enter our cost functionals. Our tracking approach is developed in greater detail in \secref{s:methods}. We define valid cell registration mappings between successive image frames in \secref{s:mapping}. We outline how to automatically  calibrate the weights of our various penalty terms in \secref{s:calibrate}. Our algorithms for pairing parent cells with their children and for cell-to-cell registration are developed in \secref{s:parent-child-paring} through \secref{s:bmmincost}. We present our main validation results on long image sequences (time series of images) in \secref{s:map_split} and conclude with \secref{s:conclusions}.

\section{Datasets}\label{s:data}

Below we introduce the datasets used to evaluate the performance of the proposed methodology. The synthetic data is described in \secref{s:synthetic}. The experimental data (real imaging data) is described in \secref{s:real-data}.

\subsection{Synthetic Videos of Simulated Cell Colonies}\label{s:synthetic}

To validate our cell tracking algorithms, we consider simulated image sequences of dense cell populations. We refer to~\cite{Winkle:2017a,Winkle:2021a} for a detailed description of this mathematical model and its implementation.\footnote{The code for generating the synthetic data has been released at \url{https://github.com/jwinkle/eQ}.} The simulated cell colony dynamics are driven by an agent based model~\cite{Winkle:2017a,Winkle:2021a}, which emulates live colonies of growing, moving, and dividing rod-like \emph{E.\ coli} cells in a 2D microfluidic trap environment. Between two successive frames $J$, $J_+$, cells are allowed to move until they nearly bump into each other, and to grow at multiplicative rate denoted $g.rate$ with an average value of $1.05$ per minute (plus/minus a small random perturbation).

\begin{table}
\caption{Benchmark datasets. To test the tracking software we consider simulated data. We have generated data of varying complexity with different interframe durations. We note that we also consider these data to train our algorithms for tracking cells. We report the label for each dataset, the interframe duration, as well as the number of frames generated. We set the \emph{cell growth factor} to $g.rate = 1.05$ per minute. We refer to the text for details about how these data have been generated.\label{t:benchmark-data}}
\centering\small
\begin{tabular}{lll}\toprule
Label  & Interframe Duration & Number of Frames \\\midrule
BENCH1 & 1 min               & 500 \\
BENCH2 & 2 min               & 300 \\
BENCH3 & 3 min               & 300 \\
BENCH6 & 6 min               & 100 \\\bottomrule
\end{tabular}
\end{table}

The cells are modeled as 2D spherocylinders of constant 1 $\mu$m width. Each cell grew exponentially in length with a doubling time of 20 minutes. To prevent division synchronization across the population when a mother cell of length $L_{\text{div}}$ divides, the two daughter cells are assigned random birth lengths $L_0(b_1) = L_1 = \delta L_{\text{div}}$ and $L_0(b_2) \eqdef L_2 = (1 -\delta)L_{\text{div}}$, where $\delta > 0$ is a random number sampled independently at each division from a uniform distribution on $[0.45, 0.55]$. Consequently, a bacterial cell $b$ of length $L_{\text{div}}$ divides into two cells $b_1$ and $b_2$, their lengths $L_1$, $L_2$ satisfy $L_1 + L_2 = L_{\text{div}}$ and $L_i /L_{\text{div}}$, $i=1,2$, is a random number. The cells have a length of approximately 2$\mu$m after division and 4$\mu$m right before division. We refer to \cite{Winkle:2021a} for additional details. The simulation keeps track of cell lineage, cell size, and cell location (among other parameters). The main output of each such simulation considered here is a binary image sequence of the cell colony with a fixed interframe duration. Each such synthetic image sequence is used as the sole input to our cell tracking algorithm. The remaining meta-data generated by the simulations are only used as ground truth to evaluate the performance of our tracking algorithms.

\begin{figure}
\centering
\begin{minipage}[c]{0.35\textwidth}
\includegraphics[totalheight=3.1cm]{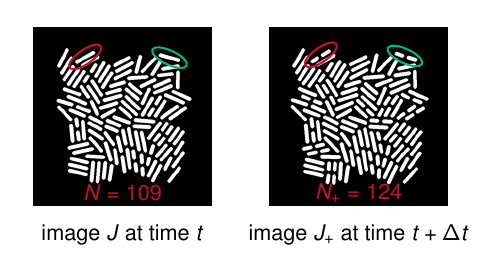}
\vspace{-0.5cm}
\end{minipage}
\begin{minipage}[c]{0.35\textwidth}
\includegraphics[totalheight=1.6cm]{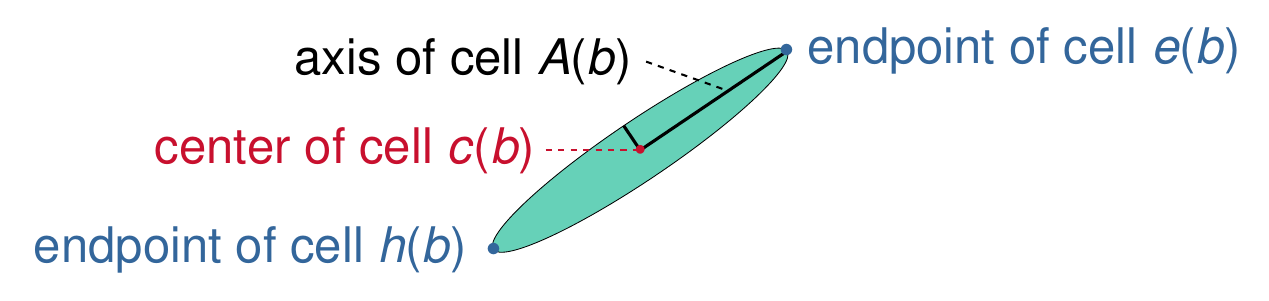}
\vspace{-0.3cm}
\end{minipage}
\caption{Simulated data and cell characteristics considered in the proposed algorithm. Left: Two successive images generated by dynamic simulation for a colony of rod-shaped bacteria. Left image $J$ displays $N = 109$ cells at time $t$. At time $t + \Delta t$ with $\Delta t = 1$\,min, cells have moved, grown, and some have divided. These cells are displayed in image $J_+$, which contains $N_+ = 124$ cells. We highlight two cells that have undergone a division between the frames (red and green ellipses). Right: Geometry of a rod shaped bacterium. We consider different quantities of interest in the proposed algorithm. These include the center $c(b)$ of a cell, the two end points $e(b)$ and $h(b)$, and the long axis $A(b)$, respectively.\label{f:syntheticdata}}
\end{figure}

We consider several several benchmark datasets of \emph{synthetic} image sequences of simulated cell colonies of different complexity. We refer to these benchmarks as BENCH1 (500 frames), BENCH2 (300 frames), BENCH3 (300 frames), and BENCH6 (100 frames), with an interframe duration of 1, 2, 3 and 6 minutes, respectively. Notice that there is no explicit noise on the growth rate. However, due to crowding of cells the growth rate will vary from cell-to-cell. The generated binary images are of size $600\times 600$ pixels. We summarize these benchmarks in \tabref{t:benchmark-data}. The associated image sequences involve between 100 up to 500 frames, respectively. In \figref{f:syntheticdata} we display an example of two simulated consecutive frames separated by 1 minute. To simplify our presentation and validation tests, we control our simulations to make sure that cells will not exit the region of interest from one frame to the next, and we exclude cells that are only partially visible in the current frames.

\subsection{Laboratory Image Sequences (Real Biological Data)}\label{s:real-data}

We also verify the performance of our approach on real datasets of \emph{E. coli} bacteria. These bacteria are about 1~$\mu$m in diameter and on average 3~$\mu$m in length, and they divide about every 30 minutes. The original images exported from the microscope are 0.11$\mu$m/pixel. The microscopy experimental data was obtained using JS006~\cite{Stricker:2008a} (BW25113 $\Delta$\emph{araC} $\Delta$lacI) \emph{E. coli} strains containing a plasmid constitutively expressing yellow or cyan fluorescent protein (\emph{sfyfp} or \emph{sfcfp}) for identification. The plasmid also contains an ampicillin resistance gene and p15A origin. These cells were grown overnight in LB medium with 100~$\mu$g/mL ampicillin for 18 hours. These cultures were diluted in the morning into 1/1000 into 50 mL fresh LB with 100~$\mu$g/ml ampicillin and grown for 3 hours until they reached an OD600 of about 0.3. The cells were then concentrated by centrifuging 30~ml of culture at 2000~g for 5 minutes and then resuspending in 10~ml of fresh LB. The concentrated culture was loaded into a hallway microfluidic device prewarmed and flushed with 0.1\% (v/v) Tween-20~\cite{Chen:2015a}. In the microfluidic device, the cells were provided with continuous fresh LB with 100~$\mu$g/ml ampicillin and 0.075\% (v/v) Tween-20. The microfluidic device was placed onto an 60$\times$ oil objective and imaged every 6 min at phase contrast, YFP, and CFP filter settings using an inverted fluorescence microscope. We show a representative dataset in \figref{f:exemplary-data}.

\subsection{Cell Characteristics}\label{s:cell_characteristics}

Next, we discuss characteristics of the \emph{E.\ coli} bacteria important for our tracking algorithm.

\textbf{Cell Geometry.} In accordance with the dynamics of bacterial colonies in microfluidic traps, the dynamic simulation software generates colonies of rod-shaped bacteria. Cell shapes can be approximated by long and thin ellipsoids, which are geometrically well identified by their center, their long axis, and the two endpoints of this long axis. The center $c(b)$ is the centroid of all pixels belonging to cell $b$. The long axis $A(b)$ of cell $b$ is computed by principal component analysis ({\bf PCA}). The endpoints $e(b)$ and $h(b)$ of cell $b$ are the first and last cell pixels nearest to $A(b)$; see \figref{f:syntheticdata} (right) for a schematic illustration.

\textbf{Cell Neighbors.} For each image frame $J$, denote $B = B(J)$ the set of fully visible cells in $J$, and by $N = N(J) = \card(B)$ the number of these cells. Let $V$ be the set of all cell centers $c(b)$ with $b \in B$. Denote $\mli{delV}$ the Delaunay triangulation~\cite{Sloan:1987a} of the finite planar set $V$ with $N$ vertices. We say that two cells $b_1$, $b_2$ in $B$ are \emph{neighbors} if they verify the following three conditions:
\begin{inparaenum}[\bf i)]
\item $(b_1,b_2)$ are connected by the edge $\mli{edg}$ of one triangle in $\mli{delV}$.
\item The edge $\mli{edg}$ does not intersect any other cell in $B$.
\item Their centers verify $\|c(b_1) - c(b_2)\| \leq \rho$, where $\rho > 0$ is a user defined parameter.
\end{inparaenum}

For the synthetic images of size $600 \times 600$ that we considered (see \secref{s:synthetic}), we take $\rho = 80$ pixels. We write $b_1 \sim b_2$ for short, whenever $b_1$, $b_2$ are neighbors (i.e, satisfy the three conditions identified above).

\textbf{Cell Motion.} Let $J$, $J_+$ denote two successive images (i.e., frames). Denote $B= B(J)$, $B_+= B(J_+)$ the associated sets of cells. Superpose temporarily the images $J$ and $J_+$ so that they then have the same center pixel. Any cell $b \in B$, which does not divide in the interframe $J \to J_+$, becomes a cell $b_+$ in image $J_+$. The \iquote{motion vector} of cell $b$ from frame $J$ to $J_+$ is then defined by $v(b) = c(b_+) - c(b)$. If the cell $b$ does divide between $J$ and $J_+$, denote $b_{\text{div}}$ the last position reached by cell $b$ at the time of cell division, and define similarly the motion $v(b) =c(b_{\text{div}}) - c(b)$. In our experimental recordings of real bacterial colonies with interframe duration 6 min, there is a \emph{fixed number} $w > 0$ such that $\|v(b)\| \leq w/2$ for all cells $b \in B(J)$ for all pairs $J$, $J_+$. In particular, we observed that for real image sequences, $w= 100$\,pixels is an adequate choice. Consequently, we select $w = 100$\,pixels for all simulated image sequences of BENCH6. For BENCH1 we select $w = 45$\,pixels, again based on a comparison with real experimental recordings. Overall, the meta-parameter $w$ is assumed to be a fixed number and to be known, since $w/2$ is an observable upper bound for the cell motion norm for a particular image sequence of a lab experiment.

\textbf{Target Window.} Recall that $J$, $J+$ are temporarily superposed. Let $U(b) \subset J_+$ be a square window of width $w$, with the same center as cell $b$. The \emph{target window} $W(b)$ is the set of all cells in $B_+$ having their centers in $U(b)$. Since $\| v(b)\| \leq w/2$,  the cell $b_+$ must belong to the target window $W(b) \subset B_+$.

\section{Methodology}\label{s:methods}

\subsection{Registration Mappings}\label{s:mapping}

Next we discuss our assumptions on a valid registration mapping that establishes cell-to-cell correspondences between two frames. Let $J$, $J_+$ denote two successive images, with cell sets $B$ and $B_+$, respectively. As above, we let $N = \card(B)$, and $N_+ = \card(B_+)$. Our goal is to track each cell from  $J$ to $J_+$. For each cell $b \in B$, there exist three possible evolutions between $J$ and $J_+$:
\begin{description}
\item[\bf Case 1:] Cell $b\in B$ did {\bf not} divide in the interframe $J \to J_+$, and has become a cell $f(b) \in B_+$; that is, $f(b)$ has grown and moved during the interframe time interval.
\item[\bf Case 2:] Cell $b\in B$ divided between $J$ and $J_+$, and generated two children cells $b_1, b_2 \in B_+$; we then denote $f(b) = (b_1,b_2) \in B_+ \times B_+$.
\item[\bf Case 3:] Cell $b\in B$ disappeared in the interframe $J \to J_+$, so that $f(b)$ is not defined.
\end{description}

To simplify our exposition, we \emph{ignore Case 3}. We discuss Case 3 in greater detail in the conclusions in \secref{s:conclusions}. Consequently, a valid (true) registration mapping $f$ will take values in the set $\{B_+\} \cup \{B_+ \times B_+\}$.

\subsection{Calibration of Cost Function Weights}\label{s:calibrate}

With the notation we introduced, fix any two finite sets $A$, $A_+$. Let $G \defeq \{g: A \to A_+\}$ be the set of all mappings $g: A \to A_+$. Fix $m$ penalty functions $\fun{pen}_k(g) \geq 0$, $k = 1, \ldots, m$. Let $g^*\in G$ be the ground truth mapping we want to discover through minimization in $g$ of some given cost function $\fun{COST}(g)$ defined by the linear combination of the penalty functions $\fun{pen}_k(g)$, the contributions of which are controlled by the cost function weights $\lambda_k >0$. In this section, we present a generic \emph{weight calibration algorithm}, extending a technique introduced and applied  in \cite{Azencott:1992a,Azencott:1994a} for Markov random fields based image analysis.

The cost function must perform well (with the same weights) for hundreds of pairs of (synthetic) images $J$, $J^+$. We consider one such synthetic pair for which the ground truth registration mapping $f \in G$ is known, and use it to compute an adequate set of weights, which will then be used on all other synthetic pairs $J$, $J^+$. Notice, that for experimental recordings of real cell colonies, no ground truth registration mappings $f$ is available. In this case, $f$ should be replaced by a set of user constructed, correct \emph{partial} mappings defined on small subsets of $A$. The proposed weight calibration algorithm will also work in those situations.

We now show how knowing one ground truth mapping $f$ can be used to derive the best feasible weights ensuring that $f$ should be a plausible minimizer of the cost functional $\fun{COST}(g)$ over $g \in G$. Let $\fun{PEN}(g) = [\fun{pen}_1(g), \ldots, \fun{pen}_m(g)]$ be the vector of $m$ penalties for any mapping $g\in G$. Let $\Lambda = [\lambda_1 , \ldots, \lambda_m]$ be the weights vectors. Then, $\fun{COST}(g) = \langle\Lambda, \fun{PEN}(g)\rangle$. Replacing $g$ by another mapping $h \neq g$ induces the penalty changes $\Delta \fun{PEN}_{g,h}  = \fun{PEN}(h) - \fun{PEN}(g)$ and the cost change $\Delta \fun{COST}(g,h) = \langle\Lambda, \Delta \fun{PEN}_{g,h} \rangle$. Now, fix any known ground truth mapping $f \in G$. We want $f$ to be a minimizer of $\fun{COST}$, so we should have
$\Delta \fun{COST}(f, f') \geq 0$ for all modifications $f \to f' \in G$.

\noindent For each $a \in A$, select an arbitrary $s(a) \in W(a)$ (where $W(a)$ is the target window for cell $a$; see \secref{s:cell_characteristics}), to define a new mapping $f' = f'_a$ from $A$ to $A_+$ by $f'_a(a) = s(a)$, and $f'_a(x) \equiv f(x)$ for all $x \neq a$. Since $f$ is a minimizer of $\fun{COST}$, this single point modification $f \to f'_a$ must generate the following cost increase
\[
\langle \Lambda, \Delta \fun{PEN}(f, f'_a) \rangle = \Delta \fun{COST}(f, f'_a) \geq 0.
\]

Denote  $V_a \in \ns{R}^m$ the vector $V_a = \Delta \fun{PEN}(f, f'_a)$. Then, the positive vector $\Lambda \in \ns{R}^m$, $\Lambda \succeq 0$, should verify the set of linear constraints $\langle \Lambda, V_a\rangle \geq 0$ for all $a \in A$. There may be too many such linear constraints. Consequently, we \emph{relax} these constraints by introducing a vector $y = [y(a)] \in \ns{R}^{\card(A)}$, $y \succeq 0$, of slack variables $y(a) \geq 0$ indexed by all the $a\in A$. (In optimization slack variables are introduced as additional unknowns to transform inequality constraints to an equality constraint and a non-negativity constraint on the slack variables.) We require the unknown positive vector $\Lambda$ and the slack variables vector $y$ to verify the system of linear constraints:
\begin{equation}\label{e:constraints}
\begin{aligned}
\langle\Lambda, V_a \rangle + y(a) & = 0 && \text{for all}\; a \in A \\
\Lambda  \succeq 0,\;\;    y & \succeq 0 && \\
\langle\Lambda, Z \rangle &\leq 1000 &&
\end{aligned}
\end{equation}

\noindent where $Z = [1,\ldots,1] \in \ns{R}^m$. The normalizing constant $1000$ can be arbitrarily changed by rescaling. We seek high positive values for $\Delta \fun{COST} (f,f'_a)$ and small $L_1$-norm for the slack variable vector $y$. So, we will seek two vectors $\Lambda \in \ns{R}^m$ and $y \in \ns{R}^{\card(A)}$ solving the following \emph{convex-concave} minimization problem, where $\gamma > 0$ is a user selected (large) meta parameter:
\begin{equation}\label{e:optimization_weights}
\minopt_{\Lambda, y} \;\; \gamma \| y \|_{L1} - \sum_{a \in A} [\langle\Lambda, V_a \rangle]^+
\end{equation}

\noindent subject to \eqref{e:constraints}, where we denote $[x]^+ \defeq \max(x, 0)$ for arbitrary $x$. To numerically solve the constrained minimization problem~\eqref{e:optimization_weights}, we use the libraries \texttt{CVXPY} and \texttt{DCCP} (disciplined convex-concave programming)~\cite{Agrawal:2018a,Diamond:2016a,Shen:2016a}. \texttt{DCCP} is a package for convex-concave programming designed to solve non-convex problems.\footnote{\texttt{DCCP} can be downloaded at \url{https://github.com/cvxgrp/dccp} (last accessed on 01/20/22).} It can handle objective functions and constraints with any known curvature as defined by the rules of disciplined convex programming~\cite{Boyd:2004a}. We give examples of numerically computed weight vectors $\Lambda$ below. The computing time was less than 30 seconds for the data that we have prepared. For simplicity, we just considered one step changes in our computations, which make the overlap penalty weak. To increase the accuracy of the model it is possible to consider a larger number of samples (i.e., multi-step changes). Note that the solutions $\Lambda$ of~\eqref{e:optimization_weights} are of course not unique, even after normalization by rescaling.

\subsection{Cell Divisions and Parent-Children Short Lineages}\label{s:parent-child-paring}

Next we discuss how we tackle the assignment problem when cells divide.

\subsubsection{Cell Divisions}

We now outline a cost function based methodology to detect cell divisions. The first step will be to seek the most likely parent for each potential pair of children cells. Fix two successive synthetic image frames $J$, $J_+$ with short interframe time equal to 1\,minute. Their cell sets $B$, $B_+$ have cardinality $N$ and $N_+$, respectively. For our synthetic image sequences, all cells $b \in B$ still exist in $B_+$---either as whole cells or after dividing into two children cells, and no new cell enters the field of view during the interframe $J \to J_+$. This forces $N_+ \geq N$, and implies that the number $\mli{DIV}$ of cell divisions occurring in this interframe verifies $\mli{DIV} = N_+ - N$. Each children pair $(b_1,b_2) \in B_+ \times B_+$ is born from a single parent $b \in B$. So the set $\mli{trueCH}$ of all such  \emph{true children pairs} must then verify
\begin{equation}\label{DIV}
\card(\mli{trueCH}) = \mli{DIV} = N_+ - N.
\end{equation}

For our video recordings of actual cell populations, during any interframe, we may have $n_{out}$ cells exiting the field of view and $n_{in}$ cells entering it, so that $|\card(\mli{trueCH}) - \mli{DIV}|$ may be of the order of $n_{in} + n_{out}$. To take this into account, we \emph{relax} the constraint in \eqref{DIV} as follows
\begin{equation}\label{DIVrel}
|\card(\mli{trueCH}) - \mli{DIV}| \leq \mli{REL},
\end{equation}

\noindent where $\mli{REL}$ is a fixed bound estimated from our experiments. For simplicity, we have restricted our methodology to the situation where $n_{in}$ and $n_{out}$ are always 0. But even in that case, there was a computational advantage to using the slightly relaxed constraint \eqref{DIVrel} with $\mli{REL}=1$.

\subsubsection{Most Likely Parent Cell for a Given Children Pair}\label{s:mostlikely}

For successive images $J$, $J_+$ with 1 minute interframe, define the set $\mli{PCH}$ of \emph{plausible children pairs} by
\begin{equation}\label{e:pcpairs}
\mli{PCH} = \{(b_1, b_2) \in B_+ \times B_+ \; \text{with centers}\;c_1, c_2\;\text{verifying}\; \| c_1 - c_2 \| < \tau\},
\end{equation}

\noindent where the threshold $\tau > 0$ is user selected and fixed for the whole benchmark set BENCH1 of synthetic image sequences.

To evaluate if a pair of cells $(b_1,b_2) \in \mli{PCH}$ can qualify as a pair of children generated by division of a parent cell $b \in B$, we now quantify the geometric distortion between $b$ and $(b_1,b_2)$. Cell division of $b$ into $b_1,b_2 \in B_+$ occurs with small motions of $b_1$, $b_2$. During the short interframe duration the initial centers $c_1$, $c_2$ of $b_1$, $b_2$ in image $J$ move by at most $w/2$ pixels each (see \secref{s:cell_characteristics}), and their initial distance to the center $c$ of $b$ is roughly at most $\|A(b)\|/4$, where $A(b)$ is the long axis of cell $b$. Hence, the centers $c$, $c_1$, $c_2$ of $b$, $b_1$, $b_2$ should verify the constraint
\begin{equation}\label{e:SHLIN}
\max \{\|c_1 - c\|, \|c_2 - c\| \} \leq w + \|A\| / 4.
\end{equation}

Define the set $\mli{SHLIN}$ of potential \emph{short lineages} as the set all triplets $(b,b_1,b_2)$ with $b\in B$, $(b_1,b_2) \in \mli{PCH}$, verifying the preceding constraint~\eqref{e:SHLIN}. For each potential lineage $(b, b_1, b_2) \in \mli{SHLIN}$, define three terms penalizing the geometric distortions between a parent $b \in B$ and a pair of children $(b_1, b_2) \in \mli{PCH}$ by the following formulas, where we denote $c$, $c_1$, $c_2$, the centers of cells $b$, $b_1$, $b_2$ and $A$, $A_1$, $A_2$ their long axes, respectively: ($i$) center distortion $\fun{cen}(b,b_1,b_2) = \| c - (c_1+c_2)/2 \|$, ($ii$) size distortion $\fun{siz}(b,b_1,b_2) = \lvert \|A\|- (\|A_1\| + \|A_2\|) \rvert$, and ($iii$) angle distortion
\[
\fun{ang}(b,b_1,b_2)
= \fun{angle}(A,A_1)
+ \fun{angle}(A,A_2)
+ \fun{angle}(A, c_2- c_1).
\]

\noindent Here, $\fun{angle}$ denotes \iquote{angles between non-oriented straight lines,} with a range from $0$ to $\pi/2$. Introduce three positive weights $\lambda_{\text{cen}}$, $\lambda_{\text{siz}}$, $\lambda_{\text{ang}}$ (to be estimated), and for every short lineage $(b,b_1,b_2) \in \mli{SHLIN}$ define its \emph{distortion cost} by
\begin{equation}\label{e:distortion}
\fun{dist}(b,b_1,b_2)
= \lambda_{\text{cen}} \fun{cen}(b,b_1,b_2)
+ \lambda_{\text{siz}} \fun{siz}(b,b_1,b_2)
+ \lambda_{\text{ang}} \fun{ang}(b,b_1,b_2).
\end{equation}

For each plausible pair of children $(b_1,b_2) \in \mli{PCH}$, we will compute the  most likely \emph{parent cell} $b^* = \fun{parent}(b_1,b_2)$ as the cell $b^* \in B$ minimizing $\fun{distortion}(b,b_1,b_2)$ in \eqref{e:distortion} over all $b \in B$, as summarized by the formula
\begin{equation}\label{e:parentcell}
b^* = \fun{parent}(b_1,b_2) = \argmin \limits_{\{b \in B \mid (b,b_1,b_2) \in \mli{SHLIN}\}} \fun{dist}(b,b_1,b_2).
\end{equation}

To force this minimization to yield a reliable estimate of $b^* = \fun{parent}(b_1,b_2)$ for most true pairs of children $(b_1,b_2)$, we calibrate the weights $\lambda_{j}$, $j \in \{\text{cen}, \text{siz}, \text{ang}\}$ by the algorithm outlined in \secref{s:calibrate}, using as \iquote{ground truth set} a fairly small set of visually identified true short lineages $(\fun{parent}, \fun{children})$. For fixed $(b_1,b_2)$, the set of potential parent cells $b \in B$ has very \emph{small size} due to the constraint \eqref{e:SHLIN}. Hence, brute force minimization of the functional $\fun{dist}(b,b_1,b_2)$ in~\eqref{e:distortion} over all $b \in B$ such that $(b,b_1,b_2) \in \mli{SHLIN}$, is a \emph{fast computation} for each $(b_1,b_2)$ in $\mli{PCH}$. The distortion minimizing $b= b^*$ yields the most likely parent cell $\fun{parent}(b_1,b_2) =b^*$. The brute force minimization in $b$ of $\fun{dist}(b,b_1,b_2)$ is still a greedy minimization in the sense that other soft constraints introduced further on are not taken into consideration during this preliminary fast computation of $b^*$.

\subsubsection{Penalties to Enforce Adequate Parent-Children Links}\label{s:parent-child}

Any true pair of children cells $\mli{pch} = (b_1, b_2)$ should belong to  $\mli{PCH}$, but  must also verify lineage and geometric constraints which we now enforce via several penalties. Note that the new penalties introduced here  are fully distinct from the three  penalties specified  above to define $\fun{dist}(b,b_1,b_2)$.

\textbf{\iquote{Lineage} Penalty.} Valid children pairs $(b_1,b_2) \in \mli{PCH}$ should be correctly matchable with their most likely parent cell $b^*= \fun{parent}(b_1,b_2)$ (see \eqref{e:parentcell}). So, we define the \iquote{lineage} penalty $\fun{lin}(b_1,b_2) = \fun{dist}(b^*,b_1,b_2)$ by
\[
\fun{lin}(b_1,b_2)
= \argmin\limits_{\{b \in b \mid (b,b_1,b_2) \in \mli{shlin}\}} \fun{dist} (b,b_1,b_2)
= \fun{dist}(\fun{parent}(b_1,b_2), b_1,b_2).
\]

\noindent Notice that the computation of $\fun{lin}(b_1,b_2)$ is quite fast.

\textbf{\iquote{Gap} Penalty.} Denote $\mli{tips}(b)$ the set of two endpoints of any cell $b$. For any pair $\mli{pch} = (b_1,b_2) \in \mli{PCH}$, define endpoints $x_1 \in \mli{tips}(b_1), x_2 \in \mli{tips}(b_2)$ and the \emph{gap} penalty $\fun{gap}(b_1,b_2)$ by
\begin{equation}\label{e:gap}
\fun{gap}(b_1,b_2)
= \|x_1 - x_2\|
= \min\{ \|x - y \| \;\text{for}\; (x,y) \!\in \!\mli{TIPS}\}
\end{equation}

\noindent with $\mli{TIPS} = \mli{tips}(b_1) \times \mli{tips}(b_2)$.

\textbf{\iquote{Dev} Penalty.} For rod-shaped bacteria, a true pair $(b_1,b_2) \in \mli{PCH}$ of just born children must have a small $\fun{gap}(b_1,b_2) = \|x_1- x_2\|$ and roughly aligned cells $b_1$ and $b_2$. For $(b_1,b_2) \in \mli{PCH}$, we quantify the \emph{deviation from  alignment} $\fun{dev}(b_1,b_2)$ as follows. Let $x_1$, $x_2$ be the closest endpoints of $b_1$, $b_2$ (see \eqref{e:gap}). Let $\mli{str}_{12}$ be the straight line linking the centers $c_1$, $c_2$ of $b_1$, $b_2$. Let $d_1$, $d_2$ be the distances from $x_1$, $x_2$ to the line $\mli{str}_{12}$. Set then
\[
\fun{dev}(b_1,b_2) = \frac{d_1 + d_2}{\|c_2 - c_1\|}.
\]

\textbf{\iquote{Ratio} Penalty.} True children pairs must have nearly equal lengths. So, for $(b_1,b_2) \in \mli{PCH}$ with lengths $L_1$, $L_2$ we define the length \emph{ratio penalty} by
\[
\fun{ratio}(b_1,b_2) = \lvert(L_1 / L_2) + (L_2 / L_1)  - 2\rvert.
\]

\textbf{\iquote{Rank} penalty.} Let $L_{\text{min}}$ be the minimum cell length over all cells in $B_+$. In $ B_+$, children pairs $(b_1,b_2)$ just born during interframe $J \to J_+$ must have lengths $L_1$, $L_2$  close to $L_{\min}$. So, for $(b_1,b_2)\in \mli{PCH}$, we define the \emph{rank} penalty by
\[
\fun{rank}(b_1,b_2) = \lvert (L_1 / L_{\text{min}}) - 1| + |(L_2 / L_{\text{min}})  - 1\rvert.
\]

Given two successive images $J$, $J_+$,  we seek the set $X = \mli{trueCH}$ of true children pairs in $B_+ \times B_+$ which is an unknown  subset  of  $\mli{PCH}$. In \secref{s:pair_matching} below we replace $X$ by its indicator function $z$ and we build
a cost function $E(z)$ which should be nearly minimized when $z$ is close to the indicator of $\mli{trueCH}$. A key term of $E(z)$ will be a weighted linear combination of the  penalty functions $\{\fun{lin}, \fun{gap}, \fun{dev}, \fun{ratio}, \fun{rank}\}$. Since these penalties are different from those introduced in \secref{s:mostlikely}, we estimate their weights in the cost function $E(z)$  by the algorithm outlined in \secref{s:calibrate}. The minimization of $E(z)$ will be implemented by simulations of a BM with energy function $E(z)$. We present these stochastic neural networks in the next section.

\subsection{Generic Boltzmann Machines (BMs)}\label{s:boltzmannmach}

Minimization of our main  cost functionals is a heavily combinatorial task, since the unknown variable is a mapping between two finite sets of sizes ranging from 80 to 120. To handle these minimizations, we use BMs originally introduced by  Hinton et al. (see~\cite{Ackley:1985a,Hinton:1986a}). Indeed, these recurrent stochastic neural networks can efficiently emulate some forms  of simulated annealing.

Each BM implemented here is a network  $\mli{BM} = \{U_1, \ldots,  U_N\}$ of $N$ \emph{stochastic neurons} $U_j$. In the BM context, the time $t = 0, 1, 2, \ldots$ is discretized and represents the number of steps in a Markov chain,  where the successive updates $Z(t) \to Z(t+1)$ of the BM configuration $Z(t)$ are analogous to the steps of a Gibbs sampler. The \emph{configuration} $Z(t) = \{Z_1(t), \ldots, Z_N(t)\}$ of the whole network $\mli{BM}$ at time $t$ is defined by  the  random states $Z_j(t)$ of all neuron $U_j$.  Each $Z_j(t)$  belongs to a fixed finite set $W(j)$. Hence $Z(t)$ belongs to the \emph{configurations set} $\mli{CONF} = W(1) \times \cdots \times W(N)$.

Neurons interactivity is specified by a finite set $\mli{CLQ}$ of \emph{cliques}. Each clique $K$ is a subset of $S= \{1, \ldots, N\}$. During  configuration updates $Z(t) \to Z(t+1)$, neurons may interact only if they are in the same clique. Here, all cliques $K$ are of small sizes 1, or 2, or 3.

For each clique $K$, one specifies an energy function $J_K(z)$ defined for all $z \in \mli{CONF}$, with $J_K(z)$ depending only on the $z_j$ such that $j \in K$. The full energy $E(z)$ of configuration $z$ is then defined by
\[
\textstyle E(z) = \sum_{K \in \mli{CLQ} }  J_K(z).
\]

The BM stochastic dynamics $Z(t) \to Z(t+1)$ is driven by the energy function $E(z)$, and by a fixed decreasing sequence of \emph{virtual temperatures} $\mli{Temp}(t) > 0$, tending slowly to $0$ as $t\to \infty$. Here we use standard temperature schemes of the form $\mli{Temp}(t) \equiv c\eta^t$ with fixed $c>0$ and slow \emph{decay rate} $0.99 < \eta < 1$.

We have implemented the classical \iquote{asynchronous} BM  dynamics. At each time $t$, \emph{only one} random neuron $U_j$ may modify its state, after reading the states of all neurons belonging to cliques containing $U_j$. A much faster alternative, implementable  on GPUs, is the \iquote{synchronous} BM  dynamics, where at each time $t$ roughly 50\% of all neurons  may simultaneously modify their states (see \cite{Azencott:1990a,Azencott:1990b,Azencott:1992c}). The detailed BM dynamics is presented in the appendix (see \secref{s:appendix}).

When the virtual temperatures  $\mli{Temp}(t)$ decrease slowly enough to $0$, the energy $E(Z(t))$ converges in probability to a local minimum of the BM energy $E(z)$ over all configurations $z \in \mli{CONF}$.

\subsection{Optimized Set of Parent-Children Triplets}\label{s:pair_matching}

Next, we formulate the search for bona fide parent-children triplets as an optimization problem. For brevity, this outline is restricted to situations where \eqref{DIV} holds, as is the case for our  synthetic image data. Simple modifications extend this approach to the relaxed constraint \eqref{DIVrel} which we used for lab videos of live cell populations. Fix successive images $J$, $J_+$ with a positive number of cell divisions $\mli{DIV}= N_+ - N$. Denote $\mli{PCH} = \{\mli{pch}_1, \mli{pch}_2, \ldots, \mli{pch}_m\}$ the set of $m$ plausible children pairs $(b_1,b_2)$ in $B_+$. The penalties $\fun{lin}$, $\fun{gap}$, $\fun{dev}$, $\fun{ratio}$, and $\fun{rank}$ defined above for all pairs $(b_1,b_2) \in \mli{PCH}$ determine five numerical vectors $\mli{LIN}$, $\mli{GAP}$, $\mli{DEV}$, $\mli{RAT}$, $\mli{RANK}$ in $\ns{R}^m$ with coordinates $\mli{LIN}_{j} = \fun{lin}(\mli{pch}_j)$, $\mli{GAP}_{j} = \fun{gap}(\mli{pch}_j)$, $\mli{DEV}_{j} = \fun{dev}(\mli{pch}_j)$, $\mli{RAT}_{j} = \fun{ratio}(\mli{pch}_j)$, $\mli{RANK}_{j}= \fun{rank}(\mli{pch}_j)$.

We now define a \emph{binary} BM constituted by $m$ \emph{binary} stochastic neurons $U_j$, $j = 1 \dots m$. At time $t = 0, 1, 2, \ldots$, each $U_j$ has a random \emph{binary valued state} $Z_j(t) =  1$ or $0$. The random configuration $Z(t) = [Z_1(t), \ldots, Z_m(t)]$ of this BM belongs to the configuration space $\mli{CONF} = \{0,1\}^m$ of all binary vectors $z = [z_1,\ldots,z_m]$. Let $\mli{SUB}$ be the set of all subsets of $\mli{PCH}$. Each configuration $z \in \mli{CONF}$ is the indicator function of a subset $\mli{sub}(z)$ of $\mli{PCH}$. We view each $\mli{sub}(z) \in \mli{SUB}$ as a possible estimate for the unknown set $\mli{trueCH} \subset B_+ \times B_+$ of true children pairs $(b_1,b_2)$. For each potential estimate $\mli{sub}(z)$ of $\mli{trueCH}$, the \iquote{lack of quality} of the estimate $\mli{sub}(z)$ will be penalized by the \emph{energy  function} $E(z) \geq 0$ of our binary BM. We now specify the energy $E(z)$ for all $z \in \mli{CONF}$ by combining the penalty terms introduced above. Note that the penalty terms introduced in \secref{s:mostlikely} are quite different from those introduced in \secref{s:parent-child}. No cell in $B_+$ can be assigned to more than one parent in $b$. To enforce this constraint, define the symmetric $m \times m$ binary matrix $[Q_{j,k}]$ by
($i$) $Q_{j,k} =1$ if $j\neq k$ and the two pairs $\mli{pch}_j$, $\mli{pch}_k$ have {\bf one} cell in common,
($ii$) $Q_{j,k} =0$ if $j\neq k$ and the two pairs $\mli{pch}_j$, $\mli{pch}_k$ have {\bf no} cell in common,
($iii$) $Q_{j,j} =0$ for all $j$.

The quadratic penalty $z \mapsto \langle z, Q z \rangle$ is non-negative for $z \in \mli{CONF}$, and must be zero if $\mli{sub}(z) = \mli{trueCH}$. Introduce six positive weight parameters to be selected further on $\lambda_j$, $j \in \{\text{lin}, \text{gap}, \text{dev}, \text{rat}, \text{rank}, Q\}$. Define the vector $V \in \ns{R}^m$ as a weighted linear combination of the penalty vectors $\mli{LIN}$, $\mli{GAP}$, $\mli{DEV}$, $\mli{RAT}$, $\mli{RANK}$
 \[
 V = \lambda_{\text{lin}} \mli{LIN}  + \lambda_{\text{gap}} \mli{GAP} + \lambda_{\text{dev}} \mli{DEV}
   + \lambda_{\text{rat}} \mli{RAT} + \lambda_{\text{rank}} \mli{RANK}.
 \]

 \noindent For any configuration $z \in \mli{CONF}$, the BM energy $E(z)$ is defined by the \emph{quadratic function}
 \begin{equation}
 \label{e:bm-energy-quad}
 E(z) = \langle V, z \rangle + \lambda_{Q} \langle z, Q z \rangle.
 \end{equation}

 \noindent We already know that the unknown set $\mli{trueCH}$ of true children pairs must have cardinal $\mli{DIV} = N_+ - N$. So we seek a configuration $z^*\in \mli{CONF}$ minimizing the energy $E(z)$ under the rigid constraint $\card \{\mli{sub}(z)\} = \mli{DIV}$. Let $\mli{ONE} \in \ns{R}^m$ be the vector with all its coordinates equal to 1. The constraint on $z$ can be reformulated as $\langle \mli{ONE}, z\rangle = \mli{DIV}$. We want the unknown $\mli{trueCH}$ to be close to the solution $z^*$ of the constrained  minimization problem
 \[
 z^* = \argmin_{z \in \mli{CONF}} E(z) \quad \text{subject to}\;\;\langle \mli{ONE}, z \rangle = \mli{DIV}.
 \]

 To force this minimization to yield a reliable estimate of $\mli{trueCH}$, we calibrate the six weights \[\lambda_j, j \in \{\text{lin}, \text{gap}, \text{dev}, \text{rat}, \text{rank}, Q\}\] by the algorithm in \secref{s:calibrate}. Denote $\mli{CONF}_1$ the set of all $z \in \mli{CONF}$ such that $\langle \mli{ONE}, z\rangle = \mli{DIV}$. To minimize $E(z)$ under the constraint $z \in \mli{CONF}_1$, fix a slowly decreasing temperature scheme $\mli{Temp}(t)$ as in \secref{s:boltzmannmach}. We need to force the BM stochastic configurations $Z(t)$ to remain in $\mli{CONF}_1$. Then, for large time step $t$, the $Z(t)$ will converge in probability  to a configuration $z^*\in \mli{CONF}_1$ approximately minimizing $E(z)$ under the constraint $z \in \mli{CONF}_1$.

 Start with any $Z(0) \in \mli{CONF}_1$. Assume that for $0 \leq s \leq t$, one has already dynamically generated BM configurations $Z(s) \in \mli{CONF}_1$. Then, randomly select two sites $j$, $k$ such that $Z_j(t) =1$ and $Z_k(t) = 0$. Compute a virtual configuration $Y$ by setting $Y_j = 0$, $Y_k = 1$, and $Y_i \equiv Z_i$ for all sites $i$ different from $j$ and $k$. Compute the energy change $\Delta E = E(Y) - E(Z(t))$, and the probability $p(t) = \exp(- D / \mli{Temp}(t))$, where $D = \max\{0, \Delta E\}$. Then randomly select $Z(t+1) = Y$ or $Z(t+1) = Z(t)$ with respective probabilities $p(t)$ and $(1-p(t))$. Clearly, this forces $Z(t+1) \in \mli{CONF}_1$.

\subsection{Performance of Automatic Children Pairing on Synthetic Videos}

In the following subsections we provide experimental  results for pairing children and parent cells.

\subsubsection{Children Pairing: Fast BM simulations}

\begin{figure}
\centering
\includegraphics[width=0.9\textwidth]{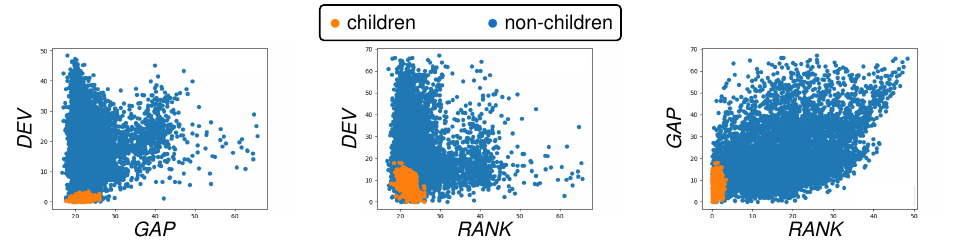}
\caption{\textit{\textbf{Scatter plots for tandems of the penalty terms $\mli{DEV}$, $\mli{GAP}$, and $\mli{RANK}$}}. We mark in orange the true children pairs and in blue invalid children pairs. These plots allow us to identify appropriate empirical thresholds to trim the (considered synthetic) data in order to reduce the computational complexity of the parent-children pairing.\label{f:pvp_children}}
\end{figure}

For $m = \card(\mli{PCH}) \leq 1000$, one can reduce the computational cost for BM dynamics simulations by pre-computing and storing the $m \times m$ symmetric binary matrix $Q$, as well as the $m$-dimensional vectors $\mli{LIN}$, $\mli{GAP}$, $\mli{DEV}$, $\mli{RAT}$, $\mli{RANK}$ and their linear combination $V$. A priori reduction of $m$ significantly reduces the computing times, and can be implemented by trimming away the pairs $\mli{pch}_j \in \mli{PCH}$ for which the penalties $\mli{LIN}_j$, $\mli{GAP}_j$, $\mli{DEV}_j$, $\mli{RAT}_j$, and $\mli{RANK}_j$ are all larger than predetermined empirical thresholds. We performed a study on 100 successive (synthetic) images. We show scatter plots for the most informative penalty terms in \figref{f:pvp_children}. These plots allow us to determine adequate thresholds for the penalty terms. We observed that for the synthetic and real data we considered the trimming of $\mli{DEV}$, $\mli{GAP}$, and $\mli{RANK}$ reduced the percentage of invalid children pairs by 95\%, therefore drastically reducing the combinatorial complexity of the problem.

The quadratic energy function $E(z)$ is the sum of clique energies $J_K(z)$ involving only cliques of cardinality 1 and 2. For any clique $K= \{j\}$ of cardinality 1, with $1 \leq j \leq m$, one has $J_K(z) = V_j z_j$. For any clique $K= \{j, k\}$ of cardinality 2, with $1 \leq j < k \leq m$, one has $J_K(z) = 2 Q_{j,k} z_j z_k$. A key computational step when generating $Z(t+1)$ is to evaluate the energy change $\Delta E$ when one flips the binary values $Z_j(t) = 1$ and $Z_k(t) = 0$ by the new value $(1 - z_i)$ for a fixed single site $i$. This step is quite fast since it uses only the numbers $V_j$, $V_k$, and $\langle q(j), Z(t) \rangle$, $\langle q(k), Z(t) \rangle$, where $q(i)$ is the $i^{th}$ row of the matrix $Q$.

\subsubsection{Children Pairing: Implementation on Synthetic Videos}

We have implemented our children pairing algorithms on synthetic image sequences having 100 to 500 image frames with 1 minutes interframe (benchmark set BENCH1; see \secref{s:synthetic}). The cell motion bound $w/2$ per interframe was defined by $w=20$ pixels. The parameter $\tau$ that defines the sets $PCH$ of plausible children pairs (see \eqref{e:pcpairs}) was set at $\tau = 45$\,pixels.

The known true cell registrations indicated that in our typical BENCH1 image sequence, the successive sets $PCH$ had average cardinals of 120, while the number of true children pairs per $\mli{PCH}$ roughly ranged from 2 to 6 with a median of 4. The size of the reduced configuration space $CONF1$ per image frame thus ranged from $10^4$ to $120^6 / 6! = 4.2\cdot10^9$ with a median of $9\cdot10^6$

Our weights estimation technique introduced in \secref{s:calibrate} yields the weights
\[
[\lambda_{\text{cen}}, \lambda_{\text{siz}}, \lambda_{\text{ang}}] = [0.255, 0.05, 0.05]
\]

\noindent and
\[
[\lambda_{\text{gap}}, \lambda_{\text{dev}}, \lambda_{\text{rat}}, \lambda_{\text{rank}}]  = [ 0.01, 1, 0.0001, 0.05]
\]

\noindent or the penalties introduced in \secref{s:map_split}. To reduce the computing time for hundreds of BM energy minimizations on the BENCH1 image sequences, we excluded obviously invalid children pairs in each $\mli{PCH}$ set, by simultaneously thresholding of the penalty terms. The BM temperature scheme was $\mli{Temp}(t) = 1000 \, (0.995)^t$, with the number of epochs capped at 5,000. The average CPU time for BM energy minimization dedicated to optimized children pairing was about $30$ seconds per frame. (We provide hardware specifications in \secref{s:hardware}.)

\subsubsection{Parent-Children Matching: Accuracy on Synthetic Videos}

For each successive image pair $J$, $J_+$, with cells $B$, $B_+$ of cardinality $N < N_+$, our parent-children matching algorithm computes a set $\mli{SHL}$ of short lineages $(b,b_1,b_2)$, where the cell $b \in B$ is expected to be the parent of cells $b_1,b_2 \in B_+$. Recall that $\mli{DIV} = N_+ - N$ provides the number of cell divisions during the interframe $J \to J_+$. The number $\mli{VAL}$ of correctly reconstructed short lineages $(b,b_1,b_2) \in \mli{SHL}$ is obtained by direct comparison to the known ground truth registration $J \to J_+$. For each frame $J$, we define the \emph{pcp-accuracy} of our Parent-Children Pairing algorithm as the ratio $\mli{VAL}/\mli{DIV}$.

We have tested our parent-children matching algorithm on three long synthetic image sequences BENCH1 (500 frames), BENCH2 (300 frames), and BENCH3 (300 frames), with respective interframes of 1, 2, and 3 minutes. For each frame $J_k$, we computed the pcp-accuracy between $J_k$ and $J_{k+1}$.

We report the accuracies of our parent-children pairing algorithms in \tabref{t:2_3_min_parents}. For BENCH1, all 500 pcp-accuracies reached 100\%. For BENCH2, pcp-accuracies reach 100\% for 298 frames out of 300, and for the remaining two frames, accuracies were still high at 93\% and 96\%. For BENCH3, where interframe duration was longest (3 minutes), the 300 pcp-accuracies decreased slightly but still averaged 99\%, and never fell below 90\%.

\begin{table}
\begin{center}
\caption{\textit{\textbf{Accuracies of  parent-children pairing algorithm}}. We applied our parent-children pairing algorithm to three long synthetic image sequences BENCH1 (500 frames), BENCH2 (300 frames), and BENCH3 (300 frames), with interframe intervals of 1, 2, 3 minutes, respectively. The table summarizes the resulting pcp-accuracies.  Note that pcp-accuracies are practically always at 100\%. For BENCH2 pcp-accuracies are 100\% for 298 frames out of 300, and for the remaining two frames, accuracies were still high at 93\% and 96\%. For BENCH3 the average pcp-accuracy for the 3 minute interframe is 99\%.}\label{t:2_3_min_parents}
\begin{tabular}{@{}lrr@{}}\toprule
\bf  sequence & \bf pcp-accuracy                 & \bf frames  \\\midrule
BENCH1        & $\mli{acc} = 100\%$              &  500 out of 500       \\
BENCH2        & $\mli{acc} = 100\%$              &  298 out of 300       \\
BENCH2        & $99\% \geq \mli{acc} \geq 93\%$  &    2 out of 300       \\
BENCH3        & $\mli{acc} = 100\%$              &  271 out of 300       \\
BENCH3        & $99\% \geq \mli{acc} \geq 95\%$  &   17 out of 300       \\
BENCH3        & $94\% \geq \mli{acc} \geq 90\%$  &   12 out of 300       \\\bottomrule
\end{tabular}
\end{center}
\end{table}

\subsection{Reduction to Registrations with No Cell Division}\label{s:map_nosplit}

Fix successive frames $J, J_+$ and their cell sets $B$, $B_+$. We seek the unknown registration mapping $f : B \to \{B_+ \cup (B_+ \times B_+)\}$, where $f(b) \in B_+$ iff cell $b$ did not divide during the interframe $J \to J_+$ and $f(b) = (b_1,b_2) \in B_+ \times B_+$ iff cell $b$ divided into $(b_1,b_2)$ during the interframe.

If $\card(B) = N < N_+ = \card(B_+)$, we know that the number of cell divisions during the interframe $J \to J_+$ should be $\mli{DIV} = \mli{DIV}(B,B_+) = N_+ - N > 0$. We then apply the parent-children matching algorithms outlined above to compute a set $\mli{SHL} = \mli{SHL}(B, B_+)$ of short lineages $(b,b_1,b_2)$ with $b \in B$, $b_1,b_2 \in B_+$ and $\card(\mli{SHL})=\mli{DIV}$. For each $(b,b_1,b_2) \in SHL$, the cell $b$ is computed by $b = \fun{parent}(b_1,b_2) $ as the parent cell of the two children cells $b_1,b_2\in B_+$.

For each $(b,b_1,b_2) \in \mli{SHL}$, eliminate from $B$ the parent cell, $b$, and eliminate from $B_+$ the two children cells $b_1$, $b_2$. We are left with two residual sets, $\mli{resB} \subset B$ and $\mli{resB}_+ \subset B_+,$ having the same cardinality, $N - \mli{DIV} = N_+ - 2 \mli{DIV}$. Assuming that our set $\mli{SHC}$ of short lineages is correctly determined, the cells $b \in \mli{redB}$ should not divide in the interframe $J \to J_+$, and hence have a single (still unknown) registration $f(b) \in \mli{redB}_+$. Thus, the still unknown part of the registration $f$ is a bijection from $\mli{redB}$ to $\mli{redB}_+$.

Let $\mli{divB} = B - \mli{redB}$ and $\mli{divB}_+= B_+ - \mli{redB}_+$. For each $b \in divB$, the cell $b$ divides into the unique pair of cells, $(b_1,b_2) \in \mli{divB}_+ \times \mli{divB}_+,$ such that $(b,b_1,b_2) \in \mli{SHL}$. Hence, we can set $f(b) = (b_1,b_2)$ for all $b \in \mli{divB}$. Thus, the remaining problem to solve is to compute the bijective registration $f : \mli{redB} \to \mli{redB}_+$. We have reduced the registration discovery to a new problem, where \emph{no cell divisions occur} in the interframe duration. In what follows, we present our algorithm to solve this registration problem.

\subsection{Automatic Cell Registration after Reduction to Cases with No Cell Division}

\begin{figure}
\centering
\includegraphics[width=0.8\textwidth]{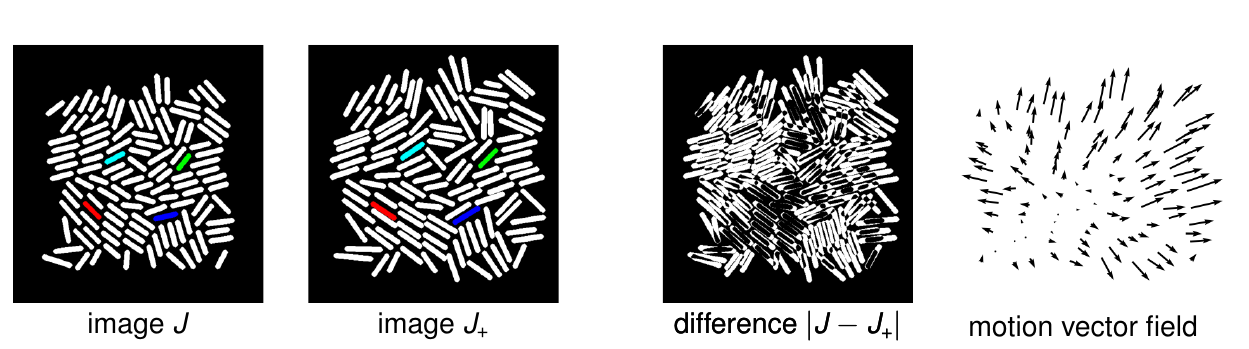}
\caption{\textit{\textbf{Simulated cell  dynamics}}. From left to right, two successive simulated images $J$ and $J_+$ with an interframe time of six minutes and no cell division, their image difference $\lvert J-J_+ \rvert$, and the associated motion vectors. For the image $J$ and $J_+$ we color four pairs of cells in $B \times B_+$, which  should be matched by the true cell registration mapping. Notice that the motion for an interframe time of six minutes is significant. We can observe that even without considering cell division, we can no longer assume that corresponding cells in frame $J$ and $J_+$ overlap.}\label{f:successive_image_nosplit}
\end{figure}

As indicated above, we can \emph{explicitly reduce} the generic cell tracking problem to a problem where there is \emph{no cell division}. We consider images $J$, $J_+$ with associated cell sets $B$, $B_+$ such that $N = \card(B) = \card(B_+)$. Hence, there are no cell divisions in the interframe $J \to J_+$ and the map $f$ of this reduced problem is (in principle) a bijection $f: B \to B_+$ with $\card(B) = \card(B_+)$. In \figref{f:successive_image_nosplit} we show two typical successive images we use for testing with no cell division generated by the simulation software~\cite{Winkle:2017a,Winkle:2021a} (see \secref{s:synthetic}).

\subsubsection{The Set $\mli{MAP}$ of Many-to-One Cell Registrations}

We have reduced the registration search to a situation where during the interframe $J \to J_+$, no cell has divided, no cell has disappeared, and no cell has suddenly emerged in $B_+$ without originating from $B$. The unknown registration $f : B \to B_+$ should then in principle be injective and onto. However, for computational efficiency, we will temporarily relax the bijectivity constraint on $f$. We will seek $f$ in the set $\mli{MAP}$ of all \emph{many-to-one mappings} $f : B \to B_+$ such that for each $b \in B$, the cell $f(b)$ is in the target window $W(b) \subset B_+$ (see \secref{s:cell_characteristics}).

\subsubsection{Registration Cost Functional}\label{s:regcostfun}

To design a cost functional $\fun{cost}(f)$, which should be roughly minimized when $f \in \mli{MAP}$ is very close to the true registration from $B$ to $B_+$, we linearly combine penalties $\fun{match}(f)$, $\fun{over}(f)$, $\fun{stab}(f)$, $\fun{flip}(f)$ weighted by unknown positive weights $\lambda_{\text{match}}$, $\lambda_{\text{over}}$, $\lambda_{\text{stab}}$, $\lambda_{\text{flip}}$, to write, for all registrations $f \in \mli{MAP}$,
\begin{equation}\label{e:costfun}
\fun{cost}(f)
= \lambda_{\text{match}} \fun{match}(f) + \lambda_{\text{over}} \fun{over}(f)
+ \lambda_{\text{stab}} \fun{stab}(f) + \lambda_{\text{flip}} \fun{flip}(f).
\end{equation}

We specify the individual terms that appear in~\eqref{e:costfun} below. Ideally, the minimizer of $\fun{cost}(f)$ over all $f \in \mli{MAP}$ is close to the unknown true registration mapping $f: B \to B_+$. To enforce a good approximation of this situation, we first estimate efficient positive weights by applying our calibration algorithm (see \secref{s:calibrate}). The actual minimization of $\fun{cost}(f)$  over all $f \in \mli{MAP}$ is then implemented by a BM described in \secref{s:bmmincost}.

\textbf{Cell Matching Likelihood: $\fun{match}(f)$.} Here, we extend a pseudo likelihood approach used to estimate parameters in Markov random fields modeling by Gibbs distributions (see \cite{Kong:2017a}). Recall that $g.\mli{rate}$ is the \emph{known} average cell growth rate. For any cells $b \in B$, $b_+ \in B_+$, the geometric quality of the matching $b \mapsto b_+$ relies on three main characteristics: ($i$) motion $c(b_+) - c(b)$ of the cell center $c(b)$, ($ii$) angle between the long axes $A(b)$ and $A(b_+)$, ($iii$) cell length ratio $\|A(b_+)\| / \|A(b)\|$. So, for all $b \in B$ and $b_+$ in the target window $W(b)$, define
\begin{inparaenum}[\bf i)]
\item Kinetic energy: $\fun{kin}(b, b_+) = \|c(b) - c(b_+)\|^2$.
\item Distortion of cell length: $\fun{dis}(b, b_+) = |\log(\|A(b_+)\|/ \|A(b)\| ) - \log g.\mli{rate}|^2$.
\item Rotation angle: $0\leq \fun{rot}(b, b_+)\leq \pi / 2$ is the geometric angle between the straight lines carrying  $A(b)$ and $A(b_+)$.
\end{inparaenum}

Fix $b \in B$, and let $b'$ run through the whole target window $W(b)$. The finite set of values thus reached by the kinetic penalties $\fun{kin}(b,b')$ has two smallest values $\mli{kin}_1(b)$, $\mli{kin}_2(b)$. Define $\mli{list}.\mli{kin} = \textstyle\bigcup_{b\in B}\{\fun{kin}_1(b), \fun{kin}_2(b)\}$, which is a list of $2 N$ \iquote{low} kinetic penalty values. Repeat this procedure for the penalties $\fun{dis}(b,b')$ and $\fun{rot}(b,b')$ to similarly define a $\mli{list}.\mli{dis}$ of $2 N$ \iquote{low} distortion penalty values, and a $\mli{list}.\mli{rot}$ of $2 N$ \iquote{low} rotation penalty values.

The three sets  $\mli{list}.\mli{kin}$, $\mli{list}.\mli{dis}$, $\mli{list}.\mli{rot}$ can be viewed as three random samples of size $2N$, respectively, generated by three unknown probability distributions $P_{\text{kin}}$, $P_{\text{dis}}$, $P_{\text{rot}}$. We approximate these three probabilities by their \emph{empirical} cumulative distribution functions $\fun{CDF}_{\text{kin}}$, $\fun{CDF}_{\text{dis}}$, $\fun{CDF}_{\text{rot}}$, which can be readily computed. We now use the right tails of these three CDFs to compute separate probabilistic evaluations of how \emph{likely} the matching of cell $b \in B$ with cell $b_+ \in W(b)$ is. For any fixed mapping $f \in \mli{MAP}$, and any $b \in B$, set $b_+ = f(b)$. Compute the three penalties $\mli{vkin} = \fun{kin}(b, b_+)$, $\mli{vdis} = \fun{dis}(b, b_+)$, $vrot = \fun{rot}(b, b_+)$, and define three associated \iquote{likelihoods} for the matching $b \to b_+ = f(b)$.
\[
\begin{aligned}
\fun{LIK}_{\text{kin}}(b,b_+) & = 1 - \fun{CDF}_{\text{kin}}(vkin),\\
\fun{LIK}_{\text{dis}}(b,b_+) & = 1 - \fun{CDF}_{\text{dis}}(vdis),\\
\fun{LIK}_{\text{rot}}(b,b_+) & = 1 - \fun{CDF}_{\text{rot}}(vrot).
\end{aligned}
\]

High values of the penalties $\mli{vkin}$, $\mli{vdis}$, $\mli{vrot}$ thus will yield  three small likelihoods for the matching $b \to b_+ = f(b)$. With this, we can define a \iquote{joint likelihood} $0 \leq \fun{LIK}(b, b_+)\leq 1$ evaluating how likely is the matching $b \to b_+ = f(b)$:
\begin{equation}\label{e:LIK}
\fun{LIK}(b, b_+) = \prod_{j \in \{\text{kin},\text{dis},\text{rot}\}} \fun{LIK}_j(b, b_+).
\end{equation}

Note that higher values of $\fun{LIK}(b, b_+)$ correspond to a better geometric quality for the matching of $b$ with $b_+ = f(b)$. To avoid vanishingly small likelihoods, whenever $\fun{LIK}(b, b_+) < 10^{-6}$, we replace it by $10^{-6}$. Then, for any mapping $f \in \mli{MAP}$, we define its  \emph{likelihood} $\fun{lik}(f)$ by the finite product
\[
\fun{lik}(f) = \prod_{b \in B} \fun{LIK}(b, f(b)).
\]

The product of these $N$ likelihoods is typically very small, since $N = \card(B)$ can be large. So, we evaluate the geometric matching quality $\fun{match}(f)$ of the mapping $f$ via the averaged \emph{log-likelihood of} $f$, namely,
\[
\fun{match}(f) = -\frac{1}{N} \log \fun{lik}(f) = - \frac{1}{N}\sum_{b\in B} \log \fun{LIK}(b,f(b)).
\]

\noindent Good registrations $f \in \mli{MAP}$ should yield small values for the criterion $\fun{match}(f)$.

\textbf{Overlap: $\fun{over}(f)$.} We expect \emph{bona fide} cell registrations  $f \in \mli{MAP}$ to be bijections. Consequently, we want to penalize mappings $f$ which are many-to-one. We say that two distinct cells $(b, b') \in B \times B$ do \emph{overlap} for the mapping $f \in \mli{MAP}$ if $f(b)= f(b')$. The total number of overlapping pairs $(b,b')$ for $f$  defines the \emph{overlap penalty}:
\[
\fun{over}(f) = \frac{1}{\card(B)} \sum_{b \in B} \sum_{b' \in B} 1_{f(b) = f(b')}.
\]

\textbf{Neighbor Stability: $\fun{stab}(f)$.} Let $B= \{b_1, \ldots, b_N\}$. Denote $G_i$ the set of all neighbors for cell $b_i$ in $B$ (i.e., $b_j \sim b_i \iff b_j \in G_i$; see \secref{s:cell_characteristics}). For \emph{bona fide} registrations $f \in \mli{MAP}$, and for most pairs of neighbors $b_i \sim b_j$ in $B$, we expect $f(b_i)$  and $f(b_j)$ to remain neighbors in $B_+$. Consequently, we penalize the lack of \iquote{neighbors stability} for $f$ by
\[
\fun{stab}(f) = \sum_i \sum_{j \neq i}\frac{1}{N \lvert G_i\rvert \lvert G_j\rvert} 1_{b_i \sim b_j} 1_{f(b_i) \not \sim f(b_j)}.
\]

\textbf{Neighbor Flip: $\fun{flip}(f)$.} Fix any mapping $f \in \mli{MAP}$, any cell $b\in B$ and any two neighbors $b'$, $b''$ of $b$ in $B$. Let $z =f(b)$, $z'= f(b')$, $z'' = f(b'')$. Let $c$, $c'$, $c''$ and $d$, $d'$, $d''$ be the centers of cells $b$, $b'$, $b''$ and $z$, $z'$, $z''$. Let $\alpha$ be the oriented angle between $c'-c$ and $c'' - c$, and let $\alpha_f$ be the angle between $d' - d$ and $d'' - d$, respectively. We say that the mapping $f$ has \emph{flipped} cells $b'$, $b''$ around $b$, and we set $\fun{FLIP}(f,b,b',b'') = 1$ if $z'$, $z''$ are both neighbors of $z$, and the two angles $\alpha$, $\alpha_f$ have \emph{opposite signs}. In all other cases, we set $\fun{FLIP}(f,b,b',b'') = 0$.

\begin{figure}
\centering
\includegraphics[width=0.48\textwidth]{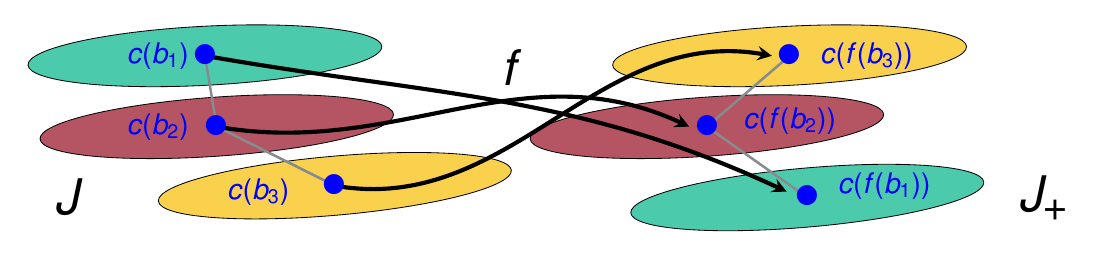}
\caption{\textit{\textbf{Illustration of an undesirable flip for the mapping $f$}}. The cells $b_1$ and $b3$ are neighbors of $b_2$, and mapped by $f$ on neighbors $z1=f(b_1), z3 =f(b3)$ of $z2 =f(b_2)$, as should be expected for bona fide cells registrations. But for this mapping $f$, we have  $z3$ above $z2$ above $z1$, whereas for the original cells we had $b_1$ above $b_2$ above $b3$. Our cost function penalizes flips of this nature.\label{f:flip}}
\end{figure}

For any registration $f \in \mli{MAP}$, define the \emph{flip penalty} for $f$ by
\[
\fun{flip}(f) \!= \!\sum_{b \in B} \sum_{b' \in B} \sum_{b'' \in B} \!\frac{1}{N \lvert G(b)\rvert^2}  \fun{FLIP}(f, b,b',b''),
\]

\noindent where $G(b)$ is the neighborhood of cell $b$ in $B$. In \figref{f:flip} we illustrate an example of an unwanted cell flip.

\subsection{BM Minimization of Registration Cost Function}\label{s:bmmincost}

In what follows, we define the optimization problem for the registration of cells from one frame to another (i.e., cell tracking), as well as associated methodology and parameter estimates.

\subsubsection{BM Minimization of $\fun{cost}(f)$ over $f \in \mli{MAP}$}

Let $B$, $B_+$ be two successive sets of cells. As outlined above, we have reduced the problem to one in which we can assume that $N = \card(B) = \card(B_+)$, so that there is no cell division during the interframe. Write $B = \{b_1, \ldots, b_N\}$. For short, denote $W(j) \subset B_+$ instead of $W(b_j)$ the target window of cell $b_j$. We seek to minimize $\fun{cost}(f)$ over all registrations $f \in \mli{MAP}$. Let $\mli{BM}$ be a BM with sites $S =\{1, \ldots, N\}$ and stochastic neurons $\{U_1, \ldots, U_N\}$. At time $t$, the random state $Z_j(t) $ of $U_j$ will be some cell $z_j$ belonging to the target window $W(j)$ and the random configuration $Z(t) =  \{Z_1(t), \ldots, Z_N(t)\}$ of the whole $\mli{BM}$ belongs to the configurations set $\mli{CONF} = W(1) \times \ldots \times W(N)$.

To any configuration $z= \{z_1, \ldots, z_N\} \in \mli{CONF}$, we associate a unique cell registration $f \in \mli{MAP}$ defined by $f(b_j) = z_j$ for all $j$, denoted by $f = \fun{map}(z)$. This determines a bijection $z \mapsto f = \fun{map}(z)$ from $\mli{CONF}$ onto $\mli{MAP}$. The inverse of $\fun{map} : \mli{CONF} \to \mli{MAP}$ will be called $\fun{range} : \mli{MAP} \to \mli{CONF}$, and is defined by $z= \fun{range}(f)$, when $z_j=f(b_j)$ for all $j$.

\subsubsection{BM Energy Function $E(z)$}

We now define the energy function $E(z) \geq 0$ of our BM for all $z \in \mli{CONF}$. Denote $E^* = \minopt_{z \in \mli{CONF}} E(z)$. Since $f \mapsto z=\fun{range}(f)$ is a bijection from $\mli{MAP}$ to $\mli{CONF}$, we must have
\[
E^* = \minopt_{z \in \mli{CONF}} E(z) = \minopt_{f \in \mli{MAP}} E(\fun{range}(f) ).
\]

Our goal is to minimize $\fun{cost}(f)$, and we know that BM simulations should roughly minimize $E(z)$ over all $z \in \mli{CONF}$. So, we define the BM energy function $E(z)$ by forcing
\begin{equation}\label{e:erange}
\fun{cost}(f) = E(\fun{range}(f))
\end{equation}

\noindent for any registration mapping $f \in \mli{MAP}$, which---due to the preceding subsection---is equivalent to
\begin{equation}\label{e:edef}
E(z) = \fun{cost}(\fun{map}(z))
\end{equation}

\noindent for all configurations $z \in \mli{CONF}$. The next subsection will explicitly express the energy $E(z)$ in terms of \emph{cliques} of neurons. Due to \eqref{e:erange} and \eqref{e:edef} we have
\[
E^* = \minopt_{f \in \mli{MAP}} \;\fun{cost}(f) = \minopt_{z \in \mli{CONF}} \;E(z).
\]

For large time $t$, the BM stochastic configuration $Z(t)$ tends with high probability to concentrate on configurations $z \in \mli{CONF}$, which roughly minimize $E(z)$. The random registration $F^t = \fun{map}(Z(t))$ will belong to $\mli{MAP}$ and verify $Z(t) = \fun{range}(F^t)$, so that $E(Z(t)) = E(\fun{range}(F^t)) = \fun{cost}(F^t))$. Consequently, for large $t$---with high probability---the random mapping $F^t = \fun{map}(Z(t))$ will have a value of the cost functional $\fun{cost}(F^t)$ close to $\minopt_{f \in \mli{MAP}} \fun{cost}(f)$.

\subsubsection{Cliques of Interactive Neurons}

The BM energy function $E(z)$ just defined turns out to involve only three sets of small cliques:
\begin{inparaenum}[\bf i)]
\item $\mli{CL}_1$ is the set of all singletons $K=\{i\}$, with $i = 1 \ldots N$.
\item $\mli{CL}_2$ is the set of all pairs $K=\{i,j\}$ such that cells $b_i$ and $b_j$ are \emph{neighbors} in $B$.
\item $\mli{CL}_3$ is the set of all triplets $K=\{i,j,k\}$ such that cells $b_j$ and $b_k$ are both \emph{neighbors} of $b_i$ in $B$.
\end{inparaenum}
Denote $\mli{CLQ} = \mli{CL}_1 \cup \mli{CL}_2 \cup \mli{CL}_3$ the set of all cliques for our BM.

\textbf{Cliques in $\mli{CL}_1$.} For each clique $K=\{i\}$ in $\mli{CL}_1$, and each $z \in \mli{CONF}$, define its energy
$J_{\text{match},K}(z) = J_{\text{match},K}(z_i)$ by
\[
J_{\text{match},K}(z) =
-\frac{1}{N} \log \fun{LIK}(b_i, z_i) \; \text{for all}\; z \in \mli{ZW},
\]

\noindent where $\fun{LIK}$ is given by \eqref{e:LIK}. Set $J_{\text{match},K} \equiv 0$ for $K$ in $\mli{CL}_2 \cup \mli{CL}_3$. For all $z \in \mli{CONF}$, define the energy $\fun{E}_{\text{match}}(z)$ by
\[
E_{\text{match}}(z)
= \!\!\!\sum_{K \in \mli{CLQ}}\!\!
J_{\text{match},K}(z)
= \!\!\!\sum_{K \in \mli{CL}_1} \!\!
J_{\text{match},K}(z),
\]

\noindent which implies that the registration $f = \fun{map}(z)$ verifies $\fun{match}(f) = E_{\text{match}}(z)$.

\textbf{Cliques in $\mli{CL}_2$.} For all $z \in \mli{CONF}$, all cliques $K = \{i, j\}$ in $\mli{CL}_2$, define the clique energies $J_{\text{over},K}(z) = J_{\text{over},K}(z_i,z_j)$ and $J_{\text{stab},K}(z) = J_{\text{stab},K}(z_i,z_j)$ by $J_{\text{over},K}(z) = 1_{z_i = z_j} / N$ and
\[
J_{\text{stab},K}(z)
= \frac{1}{N \lvert G_i\rvert \lvert G_j \rvert} 1_{b_j \sim b_i} 1_{z_j \not \sim z_i},
\]

\noindent where $\lvert G_i \rvert$ and $\lvert G_j \rvert$ are the numbers of neighbors in $B$ for cells $z_i$ and $z_j$, respectively. Set
$J_{\text{over},K} = J_{\text{stab},K} \equiv 0$
for $K$ in  $\mli{CL}_1 \cup \mli{CL}_3$. Define the two energy functions
\[
\begin{aligned}
E_{\text{over}}(z) &= \sum_{K \in \mli{CLQ}} J_{\text{over},K}(z) = \sum_{K \in \mli{CL}_2} J_{\text{over},K}(z),\\
E_{\text{stab}}(z) &= \sum_{K \in \mli{CLQ}} J_{\text{stab},K}(z) = \sum_{K \in \mli{CL}_2} J_{\text{stab},K}(z),
\end{aligned}
\]

\noindent which implies that $f = \fun{map}(z)$ verifies $\fun{over}(f) = E_{\text{over}}(z)$ and $\fun{stab}(f) = E_{\text{stab}}(z)$.

\textbf{Cliques in $\mli{CL}_3$.} For each clique $K = \{i,j,k\}$ in $\mli{CL}_3$, define the clique energy
$J_{\text{flip},K}$
by
\[
J_{\text{flip},K}(z)
= J_{\text{flip}}^{i,j,k}(z)
= \frac{1}{N \lvert G_i \rvert^2} \; \fun{FLIP}(f^{i,j,k}, b_i, b_j, b_k),
\]

\noindent where $f^{i,j,k}$ is any registration mapping $b_i$, $b_j$, $b_k$ onto $z_i$, $z_j$, $z_k$. The indicator $\fun{FLIP}$ was defined in \secref{s:regcostfun}. Set $J_{\text{flip},K} \equiv 0$ for $K$ in  $\mli{CL}_1 \cup \mli{CL}_2$. Define the energy
\[
E_{\text{flip}}(z)
= \sum_{K \in \mli{CLQ}}
J_{\text{flip},K}(z)
= \sum_{K \in \mli{CL}_3}
J_{\text{flip},K}(z),
\]

\noindent which implies that $f = F(z)$ verifies
$\fun{flip}(f)
=
E_{\text{flip}}(z)$.

Finally, define the clique energy $J_K$ for all $K \in \mli{CLQ}$ by the linear combination
\[
J_K =
  \lambda_{\text{match}} J_{\text{match},K}
+ \lambda_{\text{over}} J_{\text{over},K}
+ \lambda_{\text{stab}} J_{\text{stab},K}
+ \lambda_{\text{flip}} J_{\text{flip},K}.
\]

\noindent Summing this relation over all $K \in \mli{CLQ}$ yields
\begin{equation} \label{e:sumJK}
\sum_{K \in \mli{CLQ}} J_K
= \lambda_{\text{match}}E_{\text{match}}
+ \lambda_{\text{over}} E_{\text{over}}
+ \lambda_{\text{stab}} E_{\text{stab}}
+ \lambda_{\text{flip}} E_{\text{flip}}.
\end{equation}

\noindent Define then the final BM energy function $z \mapsto E(z)$ by
\begin{equation} \label{e:finalE(z)}
E(z) = \sum _{K \in \mli{CLQ}}  J_K(z) \;\; \text{for all} \; z\;\text{in} \; \mli{CONF}.
\end{equation}

\noindent For any $z \in \mli{CONF}$, the associated registration $f= \fun{map}(z)$ verifies $\fun{match}(f) = E_{\text{match}}(z), \fun{over}(f) = E_{\text{over}}(z)$, $\fun{stab}(f) = E_{\text{stab}}(z)$, $\fun{flip}(f) = E_{\text{flip}}(z)$. By weighted linear combination of these equalities, and due to \eqref{e:sumJK}, we obtain for all configurations $z \in \mli{CONF}$, $E(z) = \fun{cost}(f)$ when $f= \fun{map}(z)$ or, equivalently, when $z= \fun{range}(f)$.

\subsubsection{Test Set of 100 Synthetic Image Pairs}

As shown above, the minimization of $\fun{cost}(f)$ over all registrations $f \in \mli{MAP}$ is equivalent to seeking BM configurations $z \in \mli{CONF}$ with minimal energy $E(z)$. We have implemented this minimization of $E(z)$ by the long term asynchronous dynamics of the BM just defined. This algorithm was designed for the registration of image pairs exhibiting no cell division, and was, therefore, implemented after the automatic reduction of the generic registration problem, as indicated earlier. We have tested this specialized registration algorithm on a set of 100 pairs of successive images of simulated cell colonies exhibiting no cell divisions. These 100 image pairs were extracted from the benchmark set BENCH6 of synthetic image sequence described in \secref{s:synthetic}. The 100 pairs of cell sets $B$, $B_+$ had sizes $N = \card(B) = \card(B_+)$ ranging from 80 to 100 cells. For each test pair $B$, $B_+$, each target window $W(j)$ typically contained 30 to 40 cells. The set $\mli{CONF}$ of configurations had huge cardinality ranging from $10^{130}$ to $10^{160}$. But the average number of neighbors of a cell was around 4 to 5.

\subsubsection{Implementation of BM minimization for $\fun{cost}(f)$}

The numbers $\mli{clq}_1$, $\mli{clq}_2$, $\mli{clq}_3$ of cliques in $\mli{CL}_1$, $\mli{CL}_2$, $\mli{CL}_3$ have the following rough ranges $80 \leq clq1 \leq 100$, $160  \leq \mli{clq}_2 \leq 250$, and $450 \leq \mli{clq}_3 \leq  600$. For $k=1,2,3$, denote $\mli{val}(k)$ the numbers of non-zero values for $J_K(z)$ when $z$ runs through $\mli{CONF}$ and $K$ runs through all cliques of cardinality $k$. One easily checks the rough upper bounds $\mli{val}(1) < 4\,000$; $\mli{val}(2) <  200,000$;  $\mli{val}(3)  < 300,000$. Hence, to automatically register $B$ to $B_+$, one could pre-compute and store all the possible values of $J_K(z)$ for all cliques $K \in \mli{CL}_1 \cup \mli{CL}_2 \cup \mli{CL}_3$ and all the configurations $z \in \mli{CONF}$. This accelerates the key computing steps of the asynchronous BM dynamics, namely, for the evaluation of  energy change $\Delta E = E(z') - E(z)$, when configurations $z$ and  $z'$ differs at only one site $j \in S$. Indeed, the single site modification $z_j \to z'_j$ affects only the energy values $J_K(z)$ for the very small number $r(j)$ of cliques $K$, which contain the site $j$. In our benchmark sets of synthetic images, one had $r(j) < 24$ for all $j \in S$. Hence, the computation of $\Delta E$ was fast since it requires retrieving at most 24 pairs of pre-computed $J_K(z)$, $J_K(z')$, and evaluating the 24 differences $J_K(z') - J_K(z)$. Another practical acceleration step is to replace the ubiquitous computations of probabilities $p(t) = \exp(-D / \mli{Temp}(t))$ by simply testing the value $-D/\mli{Temp}(t)$ against 100 precomputed logarithmic thresholds.

In our implementation of ABM dynamics, we used virtual temperature schemes such as $\mli{Temp}(t) = 50 \cdot \rho^t$ with $0.995 \leq \rho \leq 0.999$. The BM simulation was stopped when the stochastic energy $E(Z(t))$ had remained roughly stable during the last $N$ steps. Since all target windows $W(j)$ had cardinality smaller 40, the initial configuration $Z(0) = x$ was computed via
\[
x_j= \argmax_{y \in W(j)} \fun{LIK}(b_j, y)\quad \text{for}\;j = 1, \ldots, N,
\]

\noindent where the likelihoods $\fun{LIK}$ were defined by \eqref{e:LIK}.

\subsubsection{Weight Calibration}

For the pair of successive \emph{synthetic} images $J$, $J_+$ displayed in \figref{f:successive_image_nosplit}, we have $N = \card(B) = \card(B_+) = 513$\,cells. The ground truth registration $f$ is known by construction; we used it to apply the weight calibration described in \secref{s:calibrate}. We set the meta-parameter $\gamma$ to $10^{10}$ and obtained the vector of weights
\begin{equation}\label{e:lambda*}
\Lambda^\ast
= [\lambda_{\text{match}}^\ast, \lambda_{\text{over}}^\ast, \lambda_{\text{stab}}^\ast, \lambda_{\text{flip}}^\ast]
= [110, 300, 300, 290].
\end{equation}

\noindent These weights are \emph{kept fixed for all the  100 pairs} of images taken from the set BENCH6. The determined weights are used in the cost function $\fun{cost}(f)$ defined above. This correctly parametrized the BM energy function $E(z)$. We then simulated the BM stochastic dynamics to minimize the BM energy $E(Z(t))$.

\subsubsection{BM Simulations}

We launched 100 simulations of the asynchronous BM dynamics, one for each pair of successive images in our test set of 100 images taken from BENCH6. For each such pair, the ground truth mapping $f : B \to B_+$ was known by construction and the stochastic minimization of the BM energy generated an estimated cells registration $f' : B \to B_+$. For each pair $B$, $B_+$ in the considered set of 100 images, the accuracy of this automatically computed registration $f'$ was evaluated by the percentage of cells $b \in B$ such that $f'(b) = f(b)$. When $\card(B) = N$, our BM has $N$ stochastic neurons, and the asynchronous BM dynamics proceeds by successive \emph{epochs}. Each epoch is a sequence of $N$ single site updates of the BM configuration. For each one of our 100 simulations of BM asynchronous dynamics, the number of epochs ranged from 250 to 450.

The average computing time was about eight minutes per epoch, which entailed a computing time ranging from 30 to 50 minutes for each one of our 100 automatic registrations $f' : B \to B_+$ reported here. (We specify the hardware used to carry out these computations in \secref{s:hardware}.) Each image contains about 100 to 150 cells. Consequently, the runtime for the algorithm is approximately 20 seconds per cell for our prototype implementation. We note that this is only a rough estimate. The runtime depends on several factors, such as the number of cells in an image; the number of mother and daughter cells (i.e., how many cells divide); the size of the neighborhood of each individual cell (window size); the weights used in the cost function (which affects the number of epochs); etc. We note that the temperature scheme had not been optimized yet, so that these computing times are upper bounds. Earlier SBM studies~\cite{Azencott:1993a,Azencott:1991a} indicate that the same energy minimizations on GPUs could provide a computational speedup by a factor ranging between 30 and 50. We report registration accuracies in \tabref{t:hist_one_execution_ns}. For each pair of images in the considered set of 100 images, the accuracy of automatic registration was larger than 94.5\%. The overall average registration accuracy was quite high at 99\%.

\begin{table}
\caption{\textbf{\textit{Registration accuracy for synthetic image sequence BENCH$_{100}$}}. We consider 100 pairs of consecutive synthetic images taken from the benchmark dataset BENCH6. Automatic registration was implemented by BM minimization of the cost function $\fun{cost}(f)$, which was parametrized by the vector of optimized weights $\Lambda^\ast$ in \eqref{e:lambda*}. The average registration accuracy was 99\%.\label{t:hist_one_execution_ns}}
\centering\small
\begin{tabular}{rr}\toprule
\bf registration accuracy        & \bf number of frames  \\\midrule
$acc = 100\%$                    & 55 frames out of 100  \\
$99\% \geq \mli{acc} > $ 97\%    & 40 frames out of 100  \\
$96\% \geq \mli{acc} > $ 94.5\%  & 5 frames out of 100   \\\bottomrule
\end{tabular}
\end{table}

\section{Results}\label{s:map_split}

In this section, we report results for the registration for cell dynamics involving growth, motion, and cell divisions.

\begin{figure}
\centering
\includegraphics[width=0.7\textwidth]{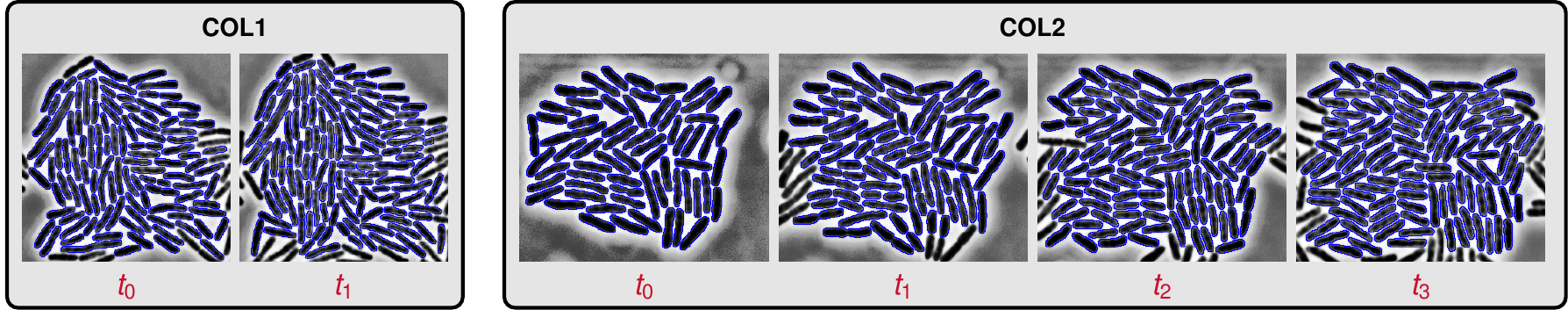}
\caption{\textbf{\textit{Segmentation results for experimental recordings of live cell colonies}}. We show two short image sequences extracts COL1 (left) and COL2 (right). The interframe duration is six minutes. The image sequence extract COL1 has only two successive image frames. The image sequence extract COL2 has four successive image frames. We are going to automatically compute four cell registrations, one for each pair of successive images in COL1 and COL2.\label{f:segmented-data}}
\end{figure}

\subsection{Tests of Cell Registration Algorithms on Synthetic Data}

We now consider more generic long synthetic image sequences of simulated cell colonies, with a small interframe duration of one minute. We still impose the mild constraint that no cell is lost between two successive images. The main difference with the earlier benchmark of 100 images from BENCH6 is that cells are \emph{allowed to freely divide} during interframes, as well as to grow and to move. For the full implementation on 100 pairs of successive images, we first execute the parent-children pairing, and remove the identified parent-children triplets; we can then apply our cell registration algorithmic on the reduced sets cells. Our image sequence contained 760 true parent-children triplets, which we automatically identified with an accuracy of 100\%. As outlined earlier, we removed all these identified cell triplets and then applied our tracking algorithm. This left us with a total of 12,631 cells (spread over 100 frames). Full automatic registration was then implemented with an accuracy higher than 99.5\%.

\subsection{Tests of Cell Registration Algorithms on Laboratory Image Sequences}

To test our cell tracking algorithm on  pairs of consecutive images extracted from recorded image sequences of bacterial colonies (real data), we had to automatically delineate all individual cells in each image. Representative frames of this data are shown in \figref{f:exemplary-data}. We describe these data in more detail in \secref{s:real-data}. We will only briefly outline the overall segmentation approach to not distract from our main contribution---the cell tracking algorithm. We use the watershed algorithm~\cite{Beucher:1979a} (also used, e.g., in \cite{Liu:2011a}) to segment each frame into individual image segments containing one single cell each. Consequently, these regions represent over segmentations of the individual cells; we only know that each region will contain a bacteria cell $b$. To segment individual cells, an additional step is necessary. We then apply adhoc nonlinear filters to remove minor segmentation artifacts. In a second step, we then identified the contour of each single cell $b$ by applying the Mumford--Shah algorithm \cite{Mumford:1989a} within the image segment containing a cell $b$. Since this procedure is quite time consuming for large images, we have implemented it to produce a training set of delineated individual cells to train a CNN for image segmentation. After automatic training, this CNN substantially reduces the runtime of the cell segmentation/delineation procedure. We show the resulting segmentations in \figref{f:segmented-data}. We provide additional information regarding our approach for the segmentation of individual bacteria cells in the appendix (see \secref{s:cell-segmentation}).

\begin{figure}
\centering
\includegraphics[width=0.7\textwidth]{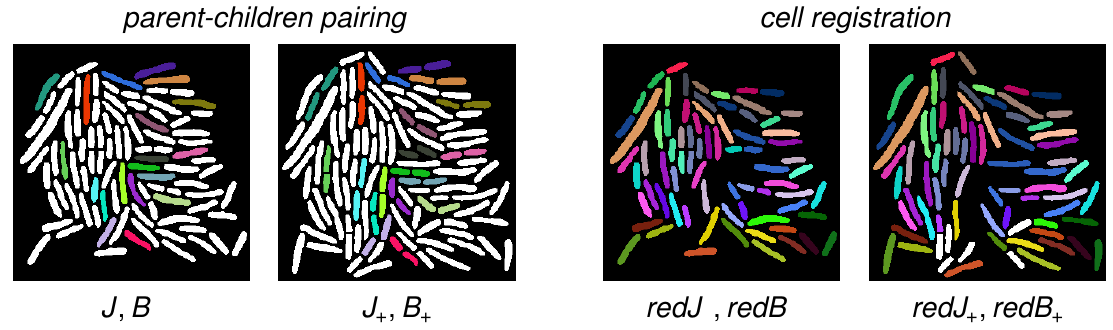}
\caption{\textbf{\textit{Cell tracking results for the pair COL1 of successive images $J$, $J_+$ shown in \figref{f:segmented-data}}}. The interframe duration is six minutes. Left: Results for parent-children pairing on COL1. Automatically detected parent-children triplets are displayed in the same color. Right: Computed registration. The removal of the automatically detected parent-children triplets (see left column) generates the reduced cell sets $\mli{redB}$ and $\mli{redB}_{+}$. Automatic registration of $\mli{redB}$ and $\mli{redB}_{+}$ is again displayed via identical color for the registered cell pairs $(b, b_+)$. Mismatches are mostly due to previous errors in parent-children pairing (see \figref{f:parent-child-paring-and-map-col2} for a more detailed assessment).\label{f:parent-child-paring-and-map-col1}}
\end{figure}

After each cell has been identified (i.e., segmented out) in each pair $J$, $J_+$ of successive images, we transform $J$, $J_+$ into binary images, where cells appear in white on a black background. For each resulting pair $B$, $B_+$ of successive sets of cells, we apply the parent-children pairing algorithm outlined in \secref{s:parent-child-paring} to identify all the short lineages. For the two successive images in COL1, the discovered short lineages are shown in \figref{f:parent-child-paring-and-map-col1} (left pair of images). Here, color designates the cell triplet algorithmically identified: parent cell in image $J$ and its two children in image $J_+$. We then remove each identified \iquote{parent} from $B$ and its two children from $B_+$. This yields the reduced cell sets $\mli{redB}$ and $\mli{redB}_+$. We can then apply our tracking algorithm (see \ref{s:map_nosplit}) dedicated to situations where cells do not divide during the interframe.

For image sequences of live cell colonies we had to re-calibrate most of our weight parameters. The weight parameters used for these image sequences are summarized in \tabref{t:weights-cost-realdata}.

\begin{table}
\caption{\textbf{\textit{Cost function weights}} for parent-children pairing in the COL1 images displayed in \figref{f:segmented-data}.\label{t:weights-cost-realdata}}
\centering\small
\begin{tabular}{lcccccccc}\toprule
Weights & $\lambda_{\text{cen}}$ & $\lambda_{\text{siz}}$ & $\lambda_{\text{ang}}$ & $\lambda_{\text{gap}}$ & $\lambda_{\text{dev}}$ &  $\lambda_{\text{rat}}$ &  $\lambda_{\text{rank}}$ &  $\lambda_{\text{over}}$ \\
Value   & $3$                    & $7$                    & $100$                  & $.8$                   & $4$                    & $.01$                   & $.01$                   & $600$                     \\\bottomrule
\end{tabular}
\end{table}

\begin{figure}
\centering
\includegraphics[width=0.7\textwidth]{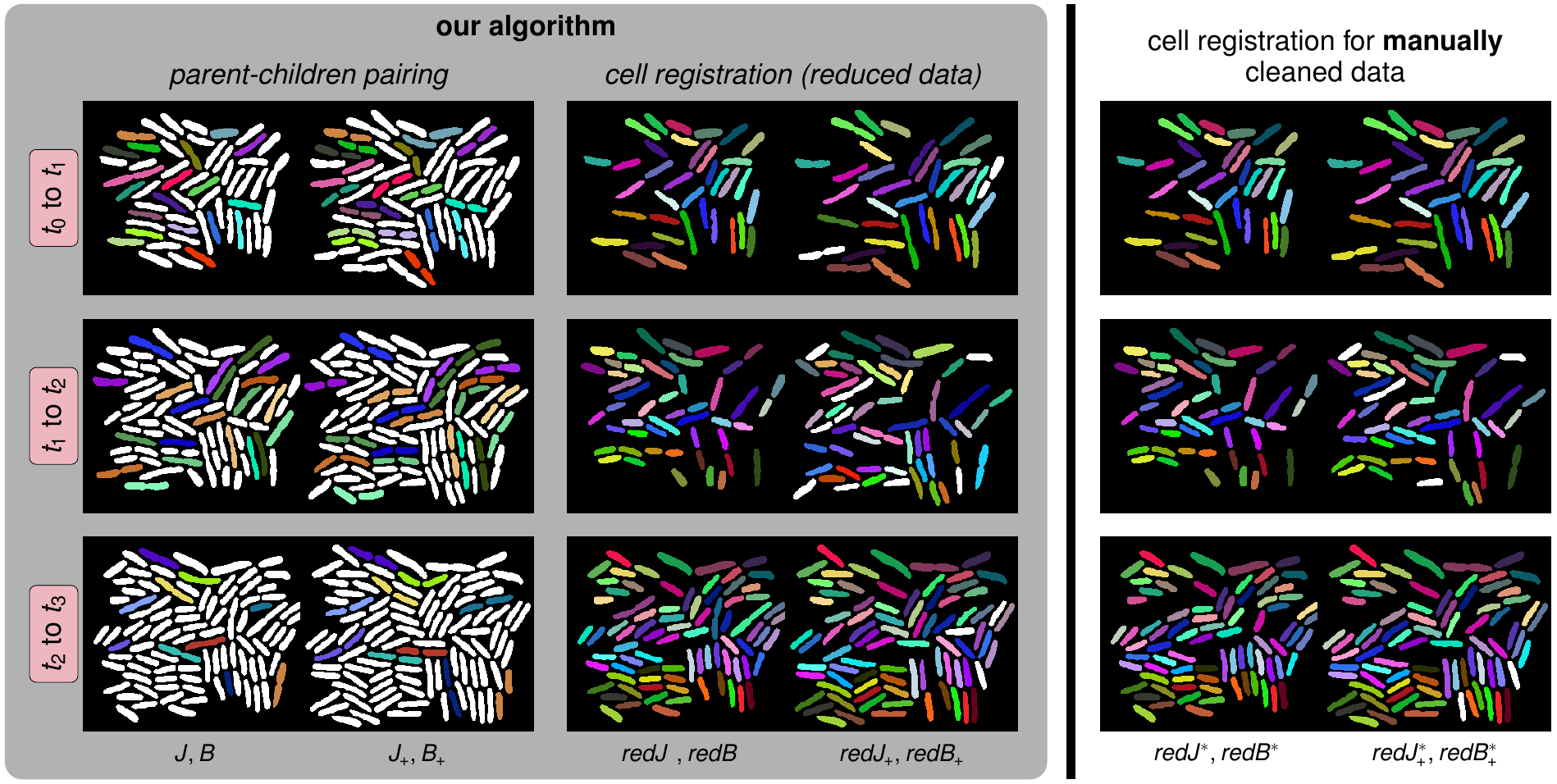}
\caption{\textbf{\textit{Cell tracking results for the short image sequence COL2 in \figref{f:segmented-data}}}. The interframe duration for COL2 is six minutes. COL2 involves four successive images $J(t_i)$, $i=0,1,2,3$. In our figure, each one of the three rows displays the automatic cell registration results between images $J(t_i) $ and $J(t_{i+1}) $ for $i=0,1,2$.  We report the accuracies of parent-children pairing and of the registration in \tabref{t:cell-matching-real-acc-col2}. Left column: Results for parent-children pairing. Each parent-children triplet is identified by the same color for each parent cell an its two children. Middle column: Display of the automatically computed registration after removing the parent-children triplets already identified in order to generate two reduced sets $\mli{redB}$ and $\mli{redB}_{+}$ of cells. Again, the same color is used for each pair of automatically registered cells. The white cells in $\mli{redB}_+$ are cells which could not be registered to some cell in  $\mli{redB}$. Right column: To differentiate between errors induced during  automatic identification of and errors generated by automatic registration between $\mli{redB}$ and $\mli{redB}_{+}$, we manually removed all \iquote{true} parent-children triplets and then applied our registration algorithm to this \iquote{cleaned} (reduced) cell sets $\mli{redB}^\ast$ and $\mli{redB}_{+}^\ast$.\label{f:parent-child-paring-and-map-col2}}
\end{figure}

\begin{table}
\caption{\textbf{\textit{Cell tracking accuracy for the short image sequence  COL2 in \figref{f:segmented-data} with an interframe of six minutes}}. We report the ratio of correctly predicted cell matches over the total number of true cell matches and the associated percentages. The accuracy  results  quantify four distinct  percentages of correct detections (i) for parent cells in image $J$, (ii) for  children cells in image $J_+$, (iii) for parent-children triplets, and (iv) for registered pairs of cells $(b,b_+) \in \mli{redB} \times \mli{redB}_+$.\label{t:cell-matching-real-acc-col2}}
\centering\small
\begin{tabular}{lrRrRrR} \toprule
\bf task                         & \multicolumn{6}{c}{\bf accuracy}  \\\midrule
                                 & \multicolumn{2}{r}{$\{t_0,t_1\}$}   & \multicolumn{2}{r}{$\{t_1,t_2\}$} & \multicolumn{2}{r}{$\{t_2,t_3\}$}  \\
correctly detected parents       &                       15/19 &  79\% &                      20/21 & 95\% &                       7/10 &  70\% \\
correctly detected children      &                       35/38 &  92\% &                      32/42 & 76\% &                      14/20 &  70\% \\
correct parent-children triplets &                       15/19 &  78\% &                      16/21 & 76\% &                       7/10 &  70\% \\
correctly registered cell pairs  &                       36/36 & 100\% &                      44/49 & 90\% &                      76/80 &  95\% \\\bottomrule
\end{tabular}
\end{table}

The BM temperature scheme was $\mli{Temp}(t) = 2000 \, (0.995)^t$, with the number of epochs capped at $5000$. We illustrate our COL1 automatic registration results in \figref{f:parent-child-paring-and-map-col1} (right pair of images). Here, if cell $b \in \mli{redB}$ has been automatically registered onto cell $b_+ \in \mli{redB}_+$, $b$, $b_+$ share the same color. The cells colored in white in $\mli{redB}_+$ are cells which the registration algorithm did not succeed in matching to some cell in $\mli{redB}$. These errors can essentially be attributed to errors in the parent-children pairing step. By visual inspection we have determined that there are 14 true parent-children triplets in the successive images of COL1. Our parent-children pairing algorithm did correctly identify 11 of these 14 triplets.To check further the performance of our registration algorithm on live images, we also report automatic registration results for \iquote{manually prepared} true versions of $\mli{redB}$ and $\mli{redB}_+$, obtained by removing \iquote{manually} the true parent-children triplets determined by visual inspection. For the short image sequence COL2, results are displayed in \figref{f:parent-child-paring-and-map-col2}.

The display setup is the same: The left column shows the results of automatic parent-children pairing. The middle column illustrates the computed registration after automatic removal of the computer identified parent-children triplets. The third column displays the computed registration after removing \iquote{manually} the true parent-children triplets determined by visual inspection. Note that the overall matching accuracy can be improved if we reduce errors in the parent-children pairing. We report quantitative accuracies in \tabref{t:cell-matching-real-acc-col2}. For parent-children pairing, accuracy ranges between 70\% and 78\%. For pure registration after correct parent-children pairing, accuracy ranges between 90\% and 100\%.

\section{Conclusions and Future Work}\label{s:conclusions}

We have developed a  methodology for automatic cell tracking in recordings of  dense bacterial colonies  growing in a mono-layer. We have also validated our approach using synthetic data from agent based simulations,  as well as experimental recordings of \emph{E.\ coli}  colonies growing in microfluidic traps. Our next goal is to streamline our implementation for systematic cell registration on experimentally acquired recordings of such cell colonies, to enable automated quantitative analysis and modeling of cell  population dynamics and lineages.

There are a number of challenges for our cell tracking algorithm: Inherent imaging artifacts such as noise or intensity drifts, cells overlaps, similarity of cell shape characteristics across the population, tight packing of cells, somewhat large interframe times, cell growth combined with cell motion and cell divisions, represent just a few of these challenges. Overall, the cell tracking problem has combinatorial complexity, and for large frames is beyond the concrete  patience of human experts. We tackle these challenges by developing a two-stage algorithm that first identifies parent-children triplets and subsequently computes cell registration from one frame to the next, after reducing the two original cell sets by automatic removal of the identified parent-children triplets. Our algorithms specify innovative cost functions dedicated to these registration challenges. These cost functions have combinatorial complexity. To discover good registrations we minimize these cost functions numerically by intensive stochastic simulations of specifically structured BMs. We have validated the potential of our approach by reporting promising results obtained on long synthetic image sequences of simulated cell colonies (which naturally provide a ground truth for cell registration from one frame to the next). We have also successfully tested our algorithms on experimental recordings of live bacterial colonies.

The choice of adequate cost functions to drive each major cost optimization step in our multi-step cell tracking algorithms is essential to obtaining good tracking. Selecting the proper formulation had a strong impact on actual tracking accuracy. Our cost functions are fundamentally nonlinear, which entails additional complications. We introduced a set of meta-parameters for each cost function, and proposed an original learning algorithm to automatically identify good ranges for these meta-parameters.

Our BMs are focused on stochastic minimization of dedicated cost functions. An interesting feature of BMs we will explore in future work is the simplicity of their natural massive parallelization for fast stochastic minimization~\cite{Azencott:1992a}. This allows us to mitigate the slow convergence typically observed for Gibbs samplers on discrete state spaces with high cardinality. Parallelized BMs implement a form of massively parallel simulated annealing. Sequential simulated annealing has been explored by physicists \cite{Mezard:1987a,Kirkpatrick:1983a,Roussel:1990a,Burda:2006a} seeking to minimize spin-glasses energies. For these clique-based energies, reaching global minima requires unfeasible CPU times, and much faster parallel simulated annealing yields only good local minima, via a sophisticated but still greedy stochastic search. Parallel stochasticity favors ending in rather stable local minima, which in turn enforces low sensitivity to small changes in energy parameters. Robustness to small changes in the coefficients of our cost functions is a desirable feature, since our algorithmic calibration of cost coefficients focuses on computing good ranges for these meta-parameters. We do not aim to seek global minima, generally a very elusive search, because computing speed and scalability are important features in our problem. Recall the established results of Huber \cite{Huber:1972a} showing that optimal estimators of the mean for a Gaussian distribution lose efficiency very quickly when the Gaussian data are slightly perturbed.

In future work, we will further improve the stability and accuracy of our cell registration algorithms by exploring natural modifications of our cost functions. In the present work, we have not yet explicitly considered the case of cells vanishing between successive frames. This is a critical issue that can occur due to cells exiting or entering the field of view as well as due to errors in cell segmentation. The problem is somewhat controlled and/or mitigated in our experimental setup, where we expect cells to enter or vanish close to a precisely positioned trap edge and/or near frame boundaries. Since we intend to track lineages, each frame-to-frame error of this type may be problematic, and it will be instrumental for our future work to address these issues.

Linking parents to children involves an optimization distinct from the final optimization of frame-to-frame registrations. This did reduce computing time without reducing the quality for our benchmark results. However, in future work one could attempt to iterate this sequence of two optimizations in order to reach a better minimum.

We note that our algorithm does work for experimental setups in which the frame rate of the video recordings is not fixed. This will require an adaptive parameter selection that depends on the frame rate. This can be implemented based on a trivial rescaling procedure. However, note that for larger interframe times, more errors will impact tracking results. Indeed, large interframe durations intensify fluctuations in key parameters of cell dynamics, and increase the range of cell displacement, imposing searches in larger cell neighborhoods for cell pairing, as well as increased combinatorial complexity.

We have considered synthetic data to evaluate the performance of our method. One clear practical issue is that some of the parameters of our tracking algorithms may change when applied to laboratory image sequences acquired from colonies of different cells, with various image acquisition setups. One can design a computational framework to automatically fit the parameters of the simulation model to the imaging data acquired on specific live cell colonies, using specific camera hardware and setup. In future work, we will attempt to implement this type of fitting for our simulation model, before launching intensive model simulations to calibrate the parameters of our new tracking algorithms. We have not yet removed physical scales in the implementation of our tracking algorithm. Implementing such a non-dimensionalization will allow us to reduce the sensitivity of our methodology with respect to new datasets.

Identification of full lineages is an interesting concrete goal for cell tracking. Evaluating the accuracy of lineage identification on real cell colonies is quite challenging since it requires inheritable biological tagging of cells. This is probably feasible for populations mixing two or three cell types, but not for individualized tagging in populations of moderate size. However even partial tagging of sub-populations would provide some control on lineage identification accuracies.

\textbf{Acknowledgements.} This work was partly supported by the National Science Foundation (NSF) through the grants DMS-1854853 (AM \& RA), DMS-2009923 (AM \& RA), 1662305 (KJ), MCB-1936770 (KJ), and DMS-2012825 (AM); the joint NSF-National Institutes of General Medical Sciences Mathematical Biology Program grant DMS-1662290 (MRB); and the Welch Foundation grant C-1729 (MRB). Any opinions, findings, and conclusions or recommendations expressed herein are those of the authors and do not necessarily reflect the views of the NSF or the Welch Foundation. This work was completed in part with resources provided by the Research Computing Data Core at the University of Houston.

\begin{appendix}

\section{Stochastic Dynamics of BMs}\label{s:appendix}

Notations and terminology refer to \secref{s:boltzmannmach}. Consider a BM  network of $N$ stochastic neurons $U_j$, with finite configuration set $\mli{CONF}= W(1) \times \ldots \times W(N)$. At time $t$, let $Z_j(t) \in W(j)$ be the random state of neuron $U_j$, and the BM configuration $Z(t) \in \mli{CONF}$ is then $Z(t) = \{ Z_1(t),\ldots,Z_N(t) \}$. Fix as in \secref{s:boltzmannmach} a sequence $\mli{Temp}(t)$ of virtual temperatures slowly decreasing to 0 for large $t$.

There are two main options to implement the Markov chain dynamics $Z(t) \to Z(t+1)$ (see \cite{Azencott:1990a}).

\subsection{Asynchronous BM  Dynamics} Generate a long random sequence of sites $m(t) \in S = \{ 1, \ldots, N \}$, for instance by concatenating successive random permutations of the set $S$. At time  $t$, the only neuron which may modify its current state is $U_{m(t)}$. For brevity, write $M= m(t)$. The neuron $U_M$ will compute its new random  state $Z_M(t+1) \in W(M)$ by the following \emph{updating procedure}:
($i$) For each $y$ in $W(M)$, define  a new  configuration $Y \in \mli{CONF}$ by $Y_M(t) = y$, and $Y_j(t) = Z_j(t)$ for all $j \neq M$. Let $\Delta(y) = E(Y) - E(Z(t))$ be the corresponding BM energy change.
($ii$) In the finite set $W(M)$, select any $z$ such that $\Delta(z) = \min_{y \in W(M)} \Delta(y)$, and set $D = \max\{0, \Delta(z)\}$.
($iii$) Compute  the probability $p = \exp(-D /\mli{Temp}(t))$.
($iv$) The new random state $Z_M(t+1)$ of neuron $U_M$ will be equal to $z$ with probability $p$ and equal to the current state $Z_M(t)$ with probability $1 - p$.
($v$) For all $j \neq M$, the new state $Z_j(t+1)$ of neuron $U_j$ remains equal to its current state $U_j(t)$.

\subsection{Synchronous BM Dynamics} Fix a \emph{synchrony} parameter $0 < \alpha < 1$, usually around 50\%. At each time $t$, \emph{all} neurons $U_j$ synchronously, but independently compute their own random \emph{binary tag}  $\mli{tag}_j(t)$, equal to 1 with probability $\alpha$, and to 0 with probability $(1 - \alpha)$. Let $\mli{SYN}(t)$ be the set of all neurons. All the neurons $U_j$ such that $\mli{tag}_j(t) = 1$ then synchronously and independently compute their new random states $Z_j(t+1) \in W(j)$ by  applying the updating procedure given above. And for all $j$ such that $\mli{tag}_j(t) = 0$, the new state $Z_j(t+1)$ of $Uj$ remains equal to $Z_j(t)$.

\subsection{Comparing Asynchronous and Synchronous BM Dynamics} As $t$ becomes large, and for temperatures $\mli{Temp}(t)$ slowly decreasing to 0,  both BM dynamics  generate with high probability configurations $Z(t)$ which provide deep local minima $E(Z(t))$ of the BM energy function. The asynchronous dynamics can be fairly slow. But the synchronous dynamics is much faster since it emulates efficient forms of \emph{parallelel simulated annealing} (see \cite{Azencott:1992a,Ram:1996a}) and is directly implementable on GPUs.

\section{Computer Hardware}\label{s:hardware}

The computations were carried out on a dedicated server at the Department of Mathematics of the University of Houston. The hardware specifications are 64 Intel(R) Xeon(R) Gold 6142 CPU cores at 2.60GHz with 128\,GB of memory.

\section{Parameters for Simulation Software}

Our tracking module is a collection of \texttt{python} functions and has been released to the public at \url{https://github.com/scopagroup/BacTrak}. We refer to~\cite{Winkle:2017a,Winkle:2021a} for a detailed description of this mathematical model and its implementation. The code for generating the synthetic data has been released at \url{https://github.com/jwinkle/eQ}. We note that detailed installation instructions for the software can be found on this page. The parameters for this agent-based simulation software are as follows: Cells were modeled as 2D spherocylinders of constant, 1 $\mu$m width. The computational framework takes into account mechanical constraints that can impact cell growth and influence other aspects of cell behavior. The growth rate of the cells is exponential and is controlled by the doubling time. The time until cells double is set to 20 min (default setting; resulting in a growth rate of $g.rate = 1.05$). The cells have a length of approximately 2$\mu$m after division and 4$\mu$m right before division (minimum division length of $4\,\mu$m; subject to some random perturbation). In our data set of simulated videos, there is no ``trap wall'' (as opposed to the simulations carried out in \cite{Winkle:2017a,Winkle:2021a}). The ``trap'' encompassing all cells on a given frame has a size of $30~\mu\text{m}\times 30~\mu\text{m}$ subdivided into $400\times 400$ pixels of size $0.075\,\mu\text{m}\times 0.075\,\mu\text{m}$. The size of the resulting binary image used in our tracking algorithm is $600\times 600$ pixels. (We add a boundary of 100 pixels on each side). Bacteria are moving, growing, dividing within the trap. However, at this stage of our study, we consider only video segments where no cell disappears and where cells do not enter the trap from outside so that the trap is a confined environment. Cells move only due to soft shocks interactions with other neighboring cells. The time interval between any two successive image frames ranges from one minute to six minutes (see \tabref{t:benchmark-data}). All other simulation parameters remain unchanged; i.e., we use the default parameters specified in the simulation software.

\section{Cell Segmentation}\label{s:cell-segmentation}

In the next couple of section we outline the framework we have developed to segment individual cells from real world laboratory imaging data. In a first step, we consider traditional segmentation algorithms---a watershed algorithm~\cite{Digabel:1978a,Beucher:1979a,Vincent:1991a} in combination with a variational contour based model---to generate a sufficiently large dataset to train a neuronal network. The actual segmentations on real data can subsequently be carried out efficiently using segmentation predictions generated by the trained neuronal network. Note that the proposed segmentation algorithm is only included for completeness. We do not view this as a major contribution of the present work.

\subsection{Watershed Algorithm}

We consider a \emph{watershed algorithm} based on immersion that compares high intensity values to local intensity minima for cell segmentation~\cite{Digabel:1978a,Beucher:1979a,Vincent:1991a}.

We consider Matlab's implementation of the watershed algorithm in the present work. This version of the watershed algorithm is unseeded and yields $n$ regions $R = \{R_1, R_2, \ldots, R_n\}$. To identify these regions, we perform a statistical analysis of each image histogram to compute adaptive rough thresholds for interiors and exterior of cells. This leads to watershed results which identify each cell by a segment slightly larger than the cell itself. The very small percentage of oversegmented cells is automatically detected by cell length and width computations through PCA analysis of each cell shape viewed as a cloud of planar points. Since our segments are slightly too wide we reduce each segment to the exact outer cell contour by applying a Mumford--Shah algorithm to each segment computed by the watershed algorithm. In an ideal case, after applying the watershed algorithm, each individual bacteria cell $b_i$, $i=1,\ldots,n$, will be located in a single region $R_i \subset \ns{R}^2$. However, we observed several segmentation errors after applying the watershed algorithm to the considered data. A common error is that a line segment that defines the boundary of a region crosses through a cell. That is, two regions contain parts of one bacterium cell. In what follows, we devise strategies to correct these errors. For this processing step we have normalized the intensities of the data to $[0,1]$.

\subsection{Segmentation Errors: Correction Steps}

We define the boundary segment $B_{i,j}$ as a non-empty intersection of two region's boundaries, i.e., $B_{i,j} = \partial S_i \cap \partial S_j$. Moreover, we denote the area of a region $R_i$ as $\fun{area}(R_i)$. We know that the interior of a bacteria cell $b_i$ has a lower intensity than the exterior region of a cell. More precisely, the interior of a cell tends to have intensity values of zero whereas the exterior of a cell (i.e., the background) tends to have an intensity that is close or equal to one. For this reason, we define a function for the intensity of the boundary. To remove outliers, we consider the average intensity value of the pixels located along a boundary segment. We denote this mean intensity value along a boundary $B_{i,j}$ by $\fun{mint}(B_{i,j})$ and the average intensity of a region $R_i$ by $\fun{mint}(R_i)$. One difficulty is that we cannot assume that the intensity of the pixels on the interior of each cell corresponds to the same value (i.e., there exist intensity and contrast drifts depending on location). We hypothesize that if $\fun{mint}(B_{i,j})$ of a boundary segment is close to the average intensity of the regions on both sides of the boundary segment $B_{i,j}$ this boundary segment does not separate two bacteria cells; it is erroneous. Conversely, if the difference between the mean intensity along a boundary segment and the mean intensity of the interior regions it separates is high, we consider that the boundary segment represents a good segmentation (i.e., represents a segment that does separate two cells). To quantify this notion, we define the height of a boundary segment as $H_{i,j} = \fun{mint}(B_{i,j}) - (\fun{mint}(R_i) + \fun{mint}(R_j))/2$.

In \tabref{t:stat_boundary}, we report some statistics associated with the quantities of interest introduced above. There are several key observations we can draw from this table which confirm our qualitative (i.e., visual) assessment of the segmentation results. Most notably, we can observe that there seem to exist outliers in terms of cell size. Moreover, we can observe that in some cases we obtain a height of the boundary segment that is negative, and by that nonsensical. These observations allow us to develop some heuristic rules to remove erroneous segmentations.

\begin{table}
\caption{Statistics of some quantities of interest related to the intensity of boundary segments and regions. These quantities allow us to define heuristics to identify erroneous segmentations computed by the watershed algorithm. We state the characteristic and report the minimum, maximum 5\% quantile, mean, and standard deviation for the reported quantities of interest.\label{t:stat_boundary}}
\centering\small
\begin{tabular}{lrrrc}\toprule
\bf Characteristic                 & \bf 5\% Quantile & \bf Min & \bf Max & \bf  Mean           \\\midrule
Watershed area                     &            56.00 &   43.00 &  984.00 & 211.00 $\pm$ 138.00 \\
Mean intensity of area             &             0.34 &    0.00 &    0.57 &   0.41 $\pm$ 0.06   \\
Mean intensity of boundary segment &             0.46 &    0.30 &    0.99 &   0.74 $\pm$ 0.14   \\
Height of boundary segment         &             0.05 &   -0.09 &    0.62 &   0.33 $\pm$ 0.14   \\\bottomrule
\end{tabular}
\end{table}

We introduce the following post-processing steps: {\bf i)} We connect small regions to their neighbors (i.e., regions that are too small in area to realistically contain any cells). We select the threshold for the area to be 65. This threshold is selected in accordance to the scale of the image and the expected size of bacteria cells observed in the image data. We merge each small region with one of its neighboring regions by removing the segment that separates the two. To select an appropriate region for merging, we choose the region that gave the lowest height $H_{i,j}$ from all available candidate regions that share the same boundary segment. {\bf ii)} We remove all boundary segments $B_{ij}$ with a height $H_{ij}$ that is below the 5\% quantile of all heights. {\bf iii)} We remove all incomplete regions from our segmentation. We define a region as incomplete, if the region or the associated boundary segments touch an edge of the image. This step is necessary since we cannot guarantee that the regions close to the boundary contain an entire cell or only parts of a cell. Consequently, we decided to remove them to prevent any issues with our post-analysis.

\subsection{Cell Boundary Detection}

The next step is to identify the boundaries of individual cells contained within a subregion defined by the watershed algorithm. To identify the boundaries of the cells (and by that segment the individual cells) we use the Mumford--Shah algorithm~\cite{Mumford:1989a}. Notice that we can execute the Mumford--Shah algorithm for each region $R_i$ separately making this an embarrassingly parallelizable problem. Denote the cell in each $R_i$ region by $b_i$. We divide each of these regions into three different zones. The first zone is the interior of the cell $b_i$ denoted by $\fun{in}(b_i)$. The second zone is exterior of the cell (i.e., the background) contained in the region and denoted by $\fun{out}(b_i)$. The third zone is the boundary of the cell $b_i$, denoted by $\partial b_i$. The Mumford--Shah algorithm represents a variational approach that allows us to segment cartoon like images. Mathematically speaking, we model information contained in each region $R_i$ as piecewise-smooth functions. In our model, the associated regions we seek to identify are given by the zones defined above---the interior and the exterior of the cell $b_i$. Let  $u_\text{int}(b_i)$ denote the mean intensity for the interior of the cell $b_i$ and $u_\text{ext}(b_i)$ denote the mean intensity for the exterior of the cell $b_i$. With this definition, we obtain the cost functional
\begin{equation*}
\fun{cost}_{\text{MS}}(\fun{int}(b_i),\fun{ext}(b_i))
= \!\!\!\sum_{x \in \fun{ext}(b_i)} \!\!\!(u(x) - u_\text{ext}(b_i))^2)
+ \!\!\!\sum_{x \in \fun{int}(b_i)} \!\!\!(u(x) - u_\text{int}(b_i))^2 + \nu\,\fun{bl}(b_i),
\end{equation*}

\noindent where the first two terms measure the discrepancy between the piecewise smooth function $u_\text{ext}$ and $u_\text{int}$ and the image intensities $u$ and the third term is a penalty that measures the length of the boundary of a particular cell $b_i$ with parameter $\nu>0$. Notice, that our formulation slightly deviates from the traditional definition of the Mumford--Shah cost functional; we drop the penalty for the smoothness of the function $u$. The minimizer of the cost function $\fun{cost}_{\text{MS}}$ defined  above provides the sought after segmentation: the boundary, interior, and exterior of a cell. We have implemented the minimization of the cost function formula for each cell separately.

\subsection{Convolutional Neural Networks (CNNs)}

Next, we introduce our actual method for cell segmentation that can be efficiently applied to a large dataset (as opposed to the prototype method described above to generate the underlying training data). The biggest issue with the methodology outlined above is that our prototype implementation is computationally costly. While we envision that an improved implementation as well as the use of parallel computing can significantly reduce the time to solution, we decided to not further pursue a reduction in runtime but extend our methodology by taking advantage of existing machine learning algorithms. Replacing the approach outlined above by CNNs allowed us to reduce the runtime by factor of 60 to less than 3 minutes, without any significant loss in accuracy.

\textbf{Training and Testing Data.} In the absence of any ground truth data set for the classification of rod-shape bacteria cells from movies of cell populations, we consider the output of the Mumford--Shah algorithm introduced above as ground truth classification for training and testing our machine learning methodology. Above, we introduced three different zones: The interior $\fun{in}(b_i)$, the exterior $\fun{ext}(b_i)$, and the boundary $\partial b_i$ of a cell $b_i$. We reduce these three regions to two zones---the interior and exterior of a cell $b_i$. We assign pixels that belong to $\fun{int}(b_i)$ the label $0$ and pixels that belong to $\fun{ext}(b_i)$ and $\partial b_i$ the label of 1. For an image of size $200 \times 200$ we obtain 40,000 binary labels. We limit the training of the CNN to a subregion of size $200\times 200$ in the center of each preprocessed image to avoid issues associated with mislabeled training data of cells located at the boundary of our data. We consider $X$ as the set of features and $Y$ as the set of labels. We want to assign to each pixel a label of either 0 or 1. For pixel $p$, we define $X_p$ to be a $7\times 7$ square window with center $p$ located in the original image. The corresponding label $Y_p$ is denoted by $C(p)$, which corresponds to the class of the pixel $p$ in the binarized image.

\textbf{CNN Algorithm.} The considered CNN algorithm consists of two parts, {\bf i)} the convolutional auto-encoder and {\bf ii)} a fully connected multiLayer perceptron (MLP). The input for the auto-encoder is a window of $7\times 7$ pixels. In the first layer of the encoder, we have a $5\times 5\times 4$ convolution layer Conv1 with $3\times 3$ kernel. We feed Conv1 to a max-pooling layer MPool2 with one stride and pooling window $2\times 2$. The output of MPool2 is the input of a $3\times 3\times 8$ convolution layer Conv3. For decoding, we have almost the same structure in reverse order: We feed Conv3 to a $5\times 5\times 4$ deconvolution with $3\times 3$ kernel. Subsequently, we feed the output of this layer to a $7\times 7\times 1$ deconvolution with $3\times 3$ kernel. The decoder's output is a window of $7\times 7$ pixels. We compare this output with the input window (since it is an auto-encoder, features and labels are the same) by using the mean square error as a cost function. We train the auto-encoder for all training sets using a mini-batch gradient descent. When the training is finished, we freeze the weights for Conv1 and Conv3.

After training the auto-encoder and freezing the weights, we feed $X$ as the input to Conv1 and get the output of Conv3 denoted by $\hat{X}$. In the next step, we train an MLP with features $\hat{X}$ and labels $Y$. We flatten $\hat{X}$, which is a $3\times 3\times 8$ matrix to a vector of size $72\times 1$, called FCL4. FCL4 is fully connected to the hidden layer HID5 with 10 nodes. We use ReLu as a nonlinear function for HID5. We connect HID5 to the output layer OUT6, which possess two nodes for the two classes 0 and 1. We use a softmax function to find two probabilistic outputs $p_0$ and $p_1 = 1- p_0$ for related classes. We use maximum-entropy as a cost function. We train the MLP for training set of $(\hat{X}, Y)$ with mini-batch gradient descent. 

We have trained the model with two images of size $200\times 200$ pixels; the training set is 80,000 $7\times 7$ images. We train the model for 100 epochs. The accuracy of the model for the image is 93\%. The confusion matrix is shown in \tabref{t:confusionML}. Based on this confusion matrix we can observe that the proposed methodology can predict the pixels located in the interior of a cell quite well. However, we can also observe that there is a slightly lower accuracy for the pixels outside the cells. This can be probably explained by the fact that the data sets are tightly packed with cells so that we have available more observations of foreground pixels (interior of cells) than pixels that belong to the background.

\begin{table}
\caption{Confusion matrix for the CNN.\label{t:confusionML}}
 \centering
\begin{tabular}{ccc}
   &    0 &    1 \\
 0 & 0.97 & 0.03 \\
 1 & 0.11 & 0.89 \\
\end{tabular}
\end{table}

\end{appendix}

\end{document}